\documentclass{ecai} 

\usepackage{latexsym}
\usepackage{amssymb}
\usepackage{amsmath}
\usepackage{amsthm}
\usepackage{booktabs}
\usepackage{enumitem}
\usepackage{graphicx}
\usepackage{color}

\usepackage{color}

\newcommand{\blue}[1]{\textcolor{blue}{#1}}

\usepackage[utf8]{inputenc} 
\usepackage[T1]{fontenc}    
\usepackage{hyperref}       
\usepackage{url}            
\usepackage{booktabs}       
\usepackage{multirow}
\usepackage{amsfonts}       
\usepackage{nicefrac}       
\usepackage{microtype}      
\usepackage{xcolor}         

\usepackage{amsfonts}
\usepackage{amsmath, amssymb}
\usepackage{amsthm}
\usepackage{relsize}
\usepackage{enumitem}
\usepackage{upgreek}
\usepackage{caption}
\usepackage{subcaption}
\usepackage{microtype}
\usepackage{graphicx}
\usepackage{booktabs} 
\usepackage{algorithm}
\usepackage{algorithmic}
\usepackage{wrapfig}
\usepackage{MnSymbol}

\newtheorem{theorem}{Theorem}

\newtheorem{proposition}{Proposition}

\newcommand{\BibTeX}{B\kern-.05em{\sc i\kern-.025em b}\kern-.08em\TeX}

\begin{document}


\begin{frontmatter}




\title{UDUC: An Uncertainty-driven Approach for Learning-based Robust Control}




\author[A]{\fnms{Yuan}~\snm{Zhang}\thanks{Corresponding Author. Email: yzhang@cs.uni-freiburg.de.}}
\author[A]{\fnms{Jasper}~\snm{Hoffman}}
\author[A]{\fnms{Joschka}~\snm{Boedecker}}

\address[A]{Neurorobotics Lab, University of Freiburg, Germany}


\begin{abstract}
    Learning-based techniques have become popular in both model predictive control (MPC) and reinforcement learning (RL). Probabilistic ensemble (PE) models offer a promising approach for modelling system dynamics, showcasing the ability to capture uncertainty and scalability in high-dimensional control scenarios. However, PE models are susceptible to mode collapse, resulting in non-robust control when faced with environments slightly different from the training set.
    In this paper, we introduce the $\textbf{u}$ncertainty-$\textbf{d}$riven rob$\textbf{u}$st $\textbf{c}$ontrol (UDUC) loss as an alternative objective for training PE models, drawing inspiration from contrastive learning. We analyze the robustness of UDUC loss through the lens of robust optimization and evaluate its performance on the challenging Real-world Reinforcement Learning (RWRL) benchmark, which involves significant environmental mismatches between the training and testing environments.
\end{abstract}

\end{frontmatter}


\section{Introduction}
\label{sec:introduction}

    Traditional control methods usually require expert knowledge of the system to design controllers. However, with the development of deep learning algorithms and easy-to-access hardware resources, learning-based methods have attracted great interest in recent years. The common approach uses interactions with the environment to learn a model, which is further used to design controllers. Such data-driven approaches are often sample-efficient and thus popular in different research fields. Examples of this are learning-based model predictive control (LBMPC)~\citep{hewing2020learningbased} in the control community and model-based reinforcement learning (MBRL) in the reinforcement learning (RL) community. 

    The objective of those learning-based methods can be briefly expressed as learning a function $f(s'|s, a)$ to fit the transition distribution $\mathcal{T}(s'|s, a)$ with the sampled dataset, where $s$ and $s'$ are states and $a$ is action. There are numerous ways to model the function $f$, e.g. parameterized physical function, Gaussian processes (GPs)~\citep{deisenroth2011pilco}, neural networks~\citep{nagabandi2018neural}. Among them, probabilistic ensemble (PE) models~\citep{chua2018deep} $f(s'|s,a) = \frac{1}{B} \sum_{b=1}^B f_b(s'|s, a)$ can capture uncertainty in the system and scale with high-dimensional data. The structure of PE is flexible to be embedded into both LBMPC~\citep{chua2018deep} and MBRL~\citep{janner2019when}, thus being a promising modelling approach to be studied.

    While PE-based methods have achieved great performance in simulated continuous control tasks (e.g. robotics), one potential concern to applying PE in real-world scenarios is its robustness. In detail, the data to train PE models usually comes from environments that are not the same as deployment. This mismatch is pervasive in control systems and may be due to (1) parameter perturbation, e.g. damping and mass of the robots could change; (2) sim-to-real mismatch, when one learns models in simulation and directly applies in reality. In addition, the trained data is finite and will inevitably suffer from sampling error. As stated in ~\citet{wang2019nonlinear}, a common concern is that the ensemble mode collapses, thus, multiple members converge to similar values, which is an undesirable case for robust control. 

    In this paper, we aim to promote the diversity of PE models for robust control inspired by the idea of contrastive learning (CL)~\citep{gutmann2010noisecontrastive}. The goal of CL is to learn such representation to effectively distinguish positive and negative samples. In the case of PE models, positive samples are collected from the training data, while negative samples are generated by other ensemble members. Intuitively, the model should learn to predict training data but be less similar to other members' predictions. We design a new loss function called $\textbf{u}$ncertainty-$\textbf{d}$riven rob$\textbf{u}$st $\textbf{c}$ontrol (UDUC) loss and explain the robustness of UDUC loss under training and testing mismatch in the view of robust optimization, based on previous study~\citet{wu2023understanding}.
    Furthermore, we evaluate UDUC on the Real-world Reinforcement Learning (RWRL) benchmark, demonstrating consistent improvements on perturbed testing environments, while only trained on one default training environment. Finally, we visualize the effects of UDUC loss to PE models on a physics-based function example. 
          

\section{Preliminaries}
\label{sec:preliminaries}

\subsection{Probabilistic Ensemble Dynamics Models}
\label{sec:ensemble_dynamics_models}

    \sloppy
    A Markov decision process (MDP)~\citep{bellman1957markovian} can be formulated as a 6-tuple $\langle \mathcal{S},\mathcal{A}, \mathcal{T}, r, \mu_0,  \gamma \rangle$. Here, $\mathcal{S}, \mathcal{A}$ represent the state and action space respectively, $\Delta_{\mathcal{S}}$ and $\Delta_{\mathcal{A}}$ the space of probability measures over $\mathcal{S}$ and $\mathcal{A}$, and $r(s,a): \mathcal{S} \times \mathcal{A} \to \mathbb{R}$\ the reward function. The initial state is sampled from an initial distribution $\mu_0: \Delta_{\mathcal{S}}$, and the future rewards are discounted by the discount factor $\gamma \in [0,1] $. The transition function $\mathcal{T}(s'|s, a): \mathcal{S} \times \mathcal{A}  \to \Delta_{\mathcal{S}} $ provides a probability distribution of the next state given the current state and action. For the real system, the exact form of the transition function is usually unknown; instead, the sampled measurements $\mathcal{D}=\{s_i, a_i, s'_{i}\}_{i=1}^N$ can be accessed from interactions. It's a common practice to learn the transition function $f: \mathcal{S} \times \mathcal{A}  \to  \Delta_{\mathcal{S}} $ by fitting the dataset $\mathcal{D}$, which can be further utilized by various model predictive control (MPC)~\citep{camacho2013model} or reinforcement learning (RL)~\citep{sutton1998reinforcement} methods, referred to popular research directions as learning-based MPC~\citep{hewing2020learningbased} and model-based RL~\citep{wang2019benchmarking}.

    There are plenty of ways to model the transition function $f$. Among them, probabilistic ensemble (PE) models~\citep{chua2018deep} can handle both aleatoric and epistemic uncertainty (due to the inherent stochasticity of the system and limited data respectively) and scale well with high-dimensional data. PE consists of $B$ probabilistic models. Each model $f_b$ is a Gaussian distribution: $f_b(s'| s, a;\theta_b)= \mathcal{N}(\mu(s, a;\theta_b), \Sigma(s, a; \theta_b))$ for its power of expression in continuous space, where $\mu: S \times A \to S$ and $\Sigma: S \times A \to S^2$ are two neural networks parameterized with $\theta_b$ to predict the mean and covariance of the next state $s'$. The final structure of a PE is $f(s'|s,a; \theta) = \frac{1}{B} \sum_{b=1}^B f_b(s'|s, a;\theta_b)$, where $\theta=\{\theta_1, \theta_2, \cdots, \theta_B\}$. Each model $f_b$ is trained on a sub-dataset $\mathcal{D}_b$, generated by independently sampling $N$ samples (with replacement) from the large dataset $\mathcal{D}$. For each sample $(s,a,s')$ in $\mathcal{D}_b$, the loss function for model $b$ is essentially the negative log-likelihood: 
    
    \begin{equation}
    \label{eq:pe_loss_function}
    \begin{aligned}
         \mathcal{L}_{\text{PE}}(\theta_b, s, a, s') &= - \log P(s'|s,a; \theta_b) \\
         &= [\mu(s, a; \theta_b) - s']^T\Sigma^{-1}(s,a;\theta_b)[\mu(s, a; \theta_b) - s'] \\
         & \quad\quad + \log \det \Sigma(s,a;\theta_b). 
    \end{aligned}
    \end{equation}
    Therefore, the final loss for model $b$ is the average over samples: 
    $\mathcal{L}(\theta_b, \mathcal{D}_b) = \frac{1}{N} \sum_{i=1}^N \mathcal{L}_{\text{PE}}(\theta_b, s_i, a_i, s_i')$. After all $B$ models have been trained, one can use the ensemble models by the trajectory sampling method~\citep{chua2018deep}. Specifically speaking, to predict the next state $s'$ at each state-action pair $(s, a)$, one model $b$ is uniformly selected from the ensemble models, and the next state $s'$ is drawn from the Gaussian distribution $f_b(s'|s, a;\theta_b)$. This propagation method is shown to best preserve the multimodality of the ensemble models, which plays a crucial role in the robustness.

\subsection{Contrastive Learning}
\label{sec:contrastive_learning}

    Contrastive learning (CL) is widely adopted in the computer vision realm~\citep{chopra2005learning, he2020momentum, chen2020simple}, aiming to gather images within the same class (positive samples) and distinguish dissimilar ones (negative samples) on an abstract embedding vector space. In this paper, we primarily focus on InfoNCE loss~\citep{oord2018representation} as the learning objective. Given a context vector $c$, one positive sample $x^+$ is drawn from the distribution $p(x|c)$, which describes the natural probability relationship between $x$ and $c$. Another $K-1$ negative samples are drawn from $q(x)$, which is a distribution independent of $c$. The positive and negative samples together form the sample set $ \mathcal{X} = \{x_1, \cdots, x_K\}$. The motivation of InfoNCE loss is to learn a representative function $g_{\theta}(x, c)$, parameterized with $\theta$, to preserve the mutual information between the sample $x$ and the context $c$, which specifically has the following form: 

    \begin{equation}
    \label{eq:infonce_loss}
    \begin{aligned}
        \mathcal{L}_{\text{InfoNCE}}(\theta, \mathcal{X}, c) = - \log \frac{\exp(g_{\theta}(x^+, c)/\tau)}{\sum_{x \in \mathcal{X}}\exp(g_{\theta}(x, c)/\tau)},
    \end{aligned}
    \end{equation}
    where the temperature $\tau$ is a key hyperparameter in InfoNCE loss to learn robust representation against sampling bias, as studied in recent work~\citep{wu2023understanding}. The optimal solution satisfies that $g_{\theta}(x, c) = \log \frac{p(x|c)}{q(x)} + \textrm{const}$, as claimed by \citet{oord2018representation}. Besides, \citet{he2020momentum} suggest using a slowly-updated target network $g_{\bar{\theta}}$ to replace the network $g_{\theta}$ in Equation~\ref{eq:infonce_loss}'s denominator for a stable learning process. As suggested in~\citep{wu2023understanding}, InfoNCE loss can be written in the expectation form as $\mathcal{L}_{\text{InfoNCE-e}}(\theta, c) = - \log \frac{ \mathbb{E}_{x^+ \sim p(x|c)} [\exp(g_{\theta}(x^+, c)/\tau)] }{\mathbb{E}_{x \sim q(x)} [\exp(g_{\theta}(x, c)/\tau)]}$, which is convenient for theoretical analysis.

\section{UDUC: Uncertainty-driven Robust Control}
\label{sec:uncertainty-driven_adaptive_control}

    \begin{figure*}[ht!]
        \centering
        \includegraphics[width=0.95\textwidth]{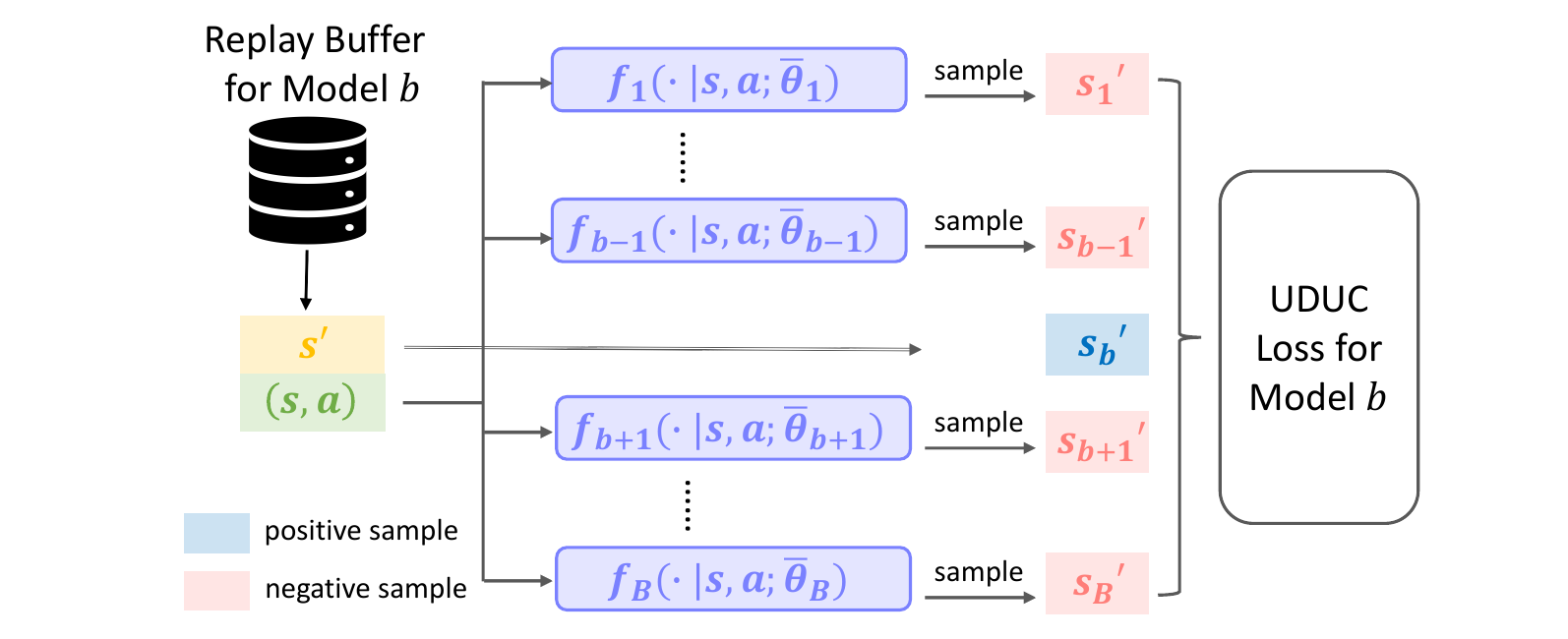}
            \caption{The overall procedure to implement UDUC loss on ensemble functions.}
        \vspace{15pt}
            \label{fig:uduc} 
    \end{figure*}

    The ensemble dynamics models introduced in Section~\ref{sec:ensemble_dynamics_models} often fail to be applied to real-world robots, and require other advanced techniques like meta reinforcement learning~\citep{nagabandi2019learning} to fine-tune the model parameters and improve the performance. The failure results from the mismatch between training and deployment environments and the vulnerable robustness of the ensemble functions. In this section, we introduce how to build on the idea of contrastive learning (CL) and improve the robustness of the ensemble functions. 
    
    Generally speaking, we enhance the standard training objectives of the probabilistic ensemble functions introduced in Section~\ref{sec:ensemble_dynamics_models} with an auxiliary CL objective. The contrastive learning objective builds on InfoNCE loss by contrasting the real next observations to the virtual next observations generated by other ensemble functions. Intuitively, this loss aims to increase the uncertainty among ensemble members' predictions and thus better apply to perturbed environments, so-called $\textbf{U}$ncertainty-$\textbf{d}$riven Rob$\textbf{u}$st $\textbf{C}$ontrol (UDUC) loss. We first introduce the specific procedures to implement UDUC loss, and further theoretically analyze it in the view of robust optimization. Finally, we provide a practical robust control approach combined with both RL-based and optimization-based controllers.

\subsection{Implementing UDUC Loss}
\label{sec:implementing_uduc_loss}

    Figure~\ref{fig:uduc} depicts the overall procedures to implement UDUC loss. As the setup in Section~\ref{sec:ensemble_dynamics_models}, there are $B$ ensemble models and each model $f_b(s'|s,a; \theta_b)$ is a Gaussian distribution on the next state $s'$. Each model maintains its own replay buffer $\mathcal{D}_b$. For a single sample $(s, a, s')$ drawn from this replay buffer, considering the current state-action pair $(s,a)$ as context $c$ and the next state $s'$ as sample $x$, one can directly construct a positive sample $s_b'$ by cloning $s'$. $B-1$ negative samples can be generated by forward passing $(s, a)$ to other $B-1$ ensemble functions (except for model $b$) and sampling the next states from these Gaussian distributions $f_i(s'|s,a; \theta_i), \forall i \neq b$. Gathering all positive and negative samples returns a sample set $\mathcal{X} = \{s_1', \cdots,  s_B'\}$. Replacing $g_{\theta}(x, c)$ in InfoNCE loss (Equation~\ref{eq:infonce_loss}) by  $-\mathcal{L}_{\text{PE}}(\theta_b, s, a, s')$, UDUC loss can be thus obtained: 

    \begin{equation}
    \label{eq:uduc_loss}
    \begin{aligned}
        &\mathcal{L}_{\text{UDUC}}(\theta_b, s, a, \mathcal{X}) = (1-\frac{1}{\tau})\mathcal{L}_{\text{PE}}(\theta_b, s, a, s_b') + \mathcal{L}_{\text{InfoNCE}}(\theta_b, s, a, \mathcal{X})\\
        &= (1-\frac{1}{\tau})\mathcal{L}_{\text{PE}}(\theta_b, s, a, s_b') -\log \frac{\exp  (-\mathcal{L}_{\text{PE}}(\theta_b, s, a, s_b')/\tau)}{\sum_{s' \in \mathcal{X}} \exp  (-\mathcal{L}_{\text{PE}}(\theta_b, s, a, s')/\tau) } \\
        &=  \mathcal{L}_{\text{PE}}(\theta_b, s, a, s_b') + \log{\sum_{s' \in \mathcal{X}} \exp(-\mathcal{L}_{\text{PE}}(\theta_b, s, a, s')/\tau)}.
    \end{aligned}
    \end{equation}

    The final loss for model $b$ is the average over samples: 
    $\mathcal{L}(\theta_b, \mathcal{D}_b) = \frac{1}{N} \sum_{i=1}^N \mathcal{L}_{\text{UDUC}}(\theta_b, s_i, a_i, \mathcal{X}_i)$. Intuitively, $\log{\sum_{s' \in \mathcal{X}} \exp(- \frac{1}{\tau} \mathcal{L}_{\text{PE}}(\theta_b, s, a, s'))}$ is the additional term apart from the standard PE loss when training the ensemble functions. This term can be interpreted as a regularizer that encourages the model $b$ to predict less similar samples to other ensemble members. Therefore, a more diverse set of transition functions is achieved and can function like domain randomization (DR) approaches~\citep{peng2018simtoreal} to trigger more robust controllers against diverse transitions. In comparison, DR randomly initializes multiple environments with various physical environmental parameters, while our method learns diverse virtual transition functions in one single training environment. This makes our method less resource-consuming and less expert knowledge required to achieve robust policies. 

    In practice, we use a target network~\citep{he2020momentum} for each model $b$ with the Gaussian distribution $f_b(s'|s,a;\bar{\theta}_b)$ to generate negative samples for the sample set $\mathcal{X}$. The target network is updated slower than the true parameter $\theta_b$ with $\bar{\theta}_b = \rho\theta_b + (1-\rho)\bar{\theta}_b$, to stabilize the learning process. Besides, when updating model $b$, the sample set $\mathcal{X}$ also includes the sample $s'$ generated from its own target network $f_b(s'|s,a;\bar{\theta}_b)$, which we call it "self-regularization". This term can be understood as the model $b$ also contrasts its past predictions generated by its target network. This essentially provides a larger set of transitions for more robust controllers, empirically shown in Section~\ref{sec:ablation_study}. 

\subsection{Understanding UDUC loss}
\label{sec:understanding_uduc_loss}
    

    In this section, we rethink UDUC loss proposed in Equation~\ref{eq:uduc_loss} with the view of robust optimization to understand its effect in detail.     
    In the case of UDUC loss, we first denote the training environment's transition distribution $\mathcal{T}_\text{train}(s'|s,a)$ ($\mathcal{T}_\text{train}$ in short), possible testing environment's transition distribution $\mathcal{T}_\text{test}(s'|s,a)$ ($\mathcal{T}_\text{test}$ in short) and probabilistic ensemble functions with target parameters $f(s'|s,a; \bar{\theta}) = \frac{1}{B} \sum_{b=1}^B f_b(s'|s,a; \bar{\theta}_b)$. When utilizing the learned ensemble models, the next state $s'$ is sampled from the distribution $f(s'|s,a;\bar{\theta})$, on which the control methods (MPC, RL) can make further plannings. The control is effective on the unseen testing environment $\mathcal{T}_\text{test}$ if the difference of the log-likelihood on these two distributions is low, i.e. $\mathbb{E}_{s' \sim f(s'|s,a;\bar{\theta})}[-\mathcal{L}_{\text{PE}}(\theta_b, s, a, s')] -  \mathbb{E}_{s' \sim \mathcal{T}_\text{test}} [-\mathcal{L}_{\text{PE}}(\theta_b, s, a, s')] \to 0$. Intuitively, this difference can be used as a regularization term of $\theta_b$ to improve its robustness in unseen environments. We formally formulate the above idea as a robust optimization problem and prove its equivalence to UDUC loss (Equation~\ref{eq:uduc_loss}) in the following Proposition.
    
    \begin{proposition}[Understanding UDUC loss]
    \label{prop:understanding_uduc_loss}
        Minimizing UDUC loss is equivalent to the following robust optimization problem:
        \begin{equation*}  
        \fontsize{8pt}{9.6pt}\selectfont
             \begin{aligned}
                &\min_{\theta_b} \mathcal{L}_{\text{RO}}(\theta_b,  s, a) = \min_{\theta_b} (1 - \frac{1}{\tau}) \underbrace{\mathbb{E}_{s' \sim \mathcal{T}_\text{train}} [\mathcal{L}_{\text{PE}}(\theta_b, s, a, s')]}_{\text{negative log-likelihood term}} \\
                 &+ \frac{1}{\tau} \underbrace{\big( \mathbb{E}_{s' \sim f(s'|s,a;\bar{\theta})} [-\mathcal{L}_{\text{PE}}(\theta_b, s, a, s')] -  \max_{\mathcal{T}_\text{test}} \mathbb{E}_{s' \sim \mathcal{T}_\text{test}} [-\mathcal{L}_{\text{PE}}(\theta_b, s, a, s')] \big)}_{\text{regularization term}}\\
                & \textit{s.t.} \quad D_{\text{KL}}(\mathcal{T}_\text{test}||\mathcal{T}_\text{train}) \le \eta, \eta \approx \mathbb{V}_{s'\sim f(s'|s,a)} [\mathcal{L}_{\text{PE}}(\theta_b, s, a, s')]/2 \tau^2.
             \end{aligned}
        \end{equation*} 
    \end{proposition}

    $\mathcal{L}_{\text{RO}}(\theta_b,  s, a)$ is the target of this robust optimization problem, consisting of a negative log-likelihood term and a regularization term. The detailed proof is built on top of \citet{wu2023understanding} and can be found in Section~\ref{sec:proof_of_proposition}. 
    Intuitively, Proposition~\ref{prop:understanding_uduc_loss} shows that UDUC loss is equivalent to a robust optimization problem with two targets. The negative log-likelihood term fits training environment $\mathcal{T}_{\text{train}}$ as the normal objective in previous work. The regularization term itself is a robust optimization objective, which aims to minimize the predictive probability ($\mathcal{L}_{\text{PE}}(\theta_b, s, a, s')$) between samples generated by the learned function $f(s'|s,a;\bar{\theta})$ and testing environment $\mathcal{T}_\text{test}$, within the neighbourhood of $\mathcal{T}_\text{train}$ (measured by the KL divergence). Accordingly, the distribution shift between these two transitions can be largely eased due to this regularization term. In addition, the temperature $\tau$ in InfoNCE loss is inversely proportional to the robust radius $\eta$, implying that smaller values of $\tau$, as an important trade-off factor, can induce larger robustness but also impact the performance on nominal training environment $\mathcal{T}_\text{train}$.
    
\subsection{Practical Robust Control Approach}
\label{sec:practical_robust_control_approach}

    The proposed UDUC loss is flexible to be plugged into both MBRL and LBMPC algorithms to improve the robustness of the controllers. Here we introduce the detailed process in Algorithm~\ref{alg:uduc_algorithm}. Line 7 to 12 denote the model update process, which essentially utilizes UDUC loss to learn parameters $\theta$. Line 13 to 15 update the agent with the model dataset $\mathcal{D}_\text{model}$ generated by transition function $f(s'|s,a; \bar{\theta)}$. Notably, this step can be omitted for MPC methods (e.g. CEM~\citep{rubinstein1999crossentropy}) since it directly utilizes transition function $f(s'|s,a; \bar{\theta)}$ for control. For neural-network-policy $\pi(a|s)$ in RL, any standard on-policy and off-policy algorithms (e.g. SAC~\citep{haarnoja2018soft}, PPO~\citep{schulman2017proximal}) can be adopted to improve the policy. Notably, both MBRL and LBMPC necessitate the utilization of transition function $f(s'|s,a; \bar{\theta)}$ to rollout the trajectory. As introduced in Section~\ref{sec:ensemble_dynamics_models}, we adopt the trajectory sampling approach by randomly selecting one model $b$ among all ensembles at each step and sampling next state $s'$ from the Gaussian distribution $f(s'|s,a; \bar{\theta}_b)$. The trajectory sampling approach is considered to maintain the multimodality of the trajectory~\citep{janner2019when}, which is necessary to generate diverse samples as motivated by UDUC loss. The detailed procedures to construct control policies with probabilistic ensemble functions for both MPC and RL are explained in Appendix~\ref{sec:control_with_probabilistic_ensemble_functions}.
    
    \begin{algorithm}[ht!]
        \caption{UDUC Algorithm}
        \label{alg:uduc_algorithm}
            \begin{algorithmic}[1]
                \STATE \textbf{Input:} ensemble transition function $f(s'|s, a;\theta)$ with parameters $\theta=\{\theta_1, \cdots, \theta_B\}$, policy function $\pi(a|s)$, environment dataset $\mathcal{D}_\text{env}=\{\}$, model dataset $\mathcal{D}_\text{model}=\{\}$, target update rate $\rho$, initial state $s$, model-agent update frequency $F_\text{model}$ and $F_\text{agent}$
                \STATE Set target transition function's parameters $\bar{\theta} \leftarrow \theta$ 
                \FOR{steps $i = 1,2,...$}
                    \STATE Execute $a \sim \pi(a|s)$  in the training environment 
                    \STATE Observe reward $r$ and next state $s'$ \\
                    \STATE $\mathcal{D}_\text{env} \leftarrow \mathcal{D}_\text{env}  \cup \{ s, a, r, s'\} $ and $s \leftarrow s'$
                    \IF{$i \mod F_\text{model} \equiv 0$ \blue{ /*update model*/}}  
                        \STATE Update parameters $\theta$ by minimizing Equation~\ref{eq:uduc_loss}
                        \STATE Update target parameters $\bar{\theta} \leftarrow (1-\rho)\bar{\theta} + \rho \theta $ 
                        \STATE Rollout transitions $(s_n, a_n, r_n, s_{n+1})_{n=1}^N$ with $f(s'|s,a; \bar{\theta})$, $\pi(a|s)$ and  $r(s, a)$\footnote{Notably, for unknown reward $r(s, a)$, it can be learned by incorporating into transition  $f(s'|s,a)$~\citep{janner2019when}.}
                        \STATE Store in the model dataset $\mathcal{D}_\text{model} \leftarrow \mathcal{D}_\text{model} \cup \{ (s_n, a_n, r_n, s_{n+1})_{n=1}^N\}$
                    \ENDIF
                    \IF{$i \mod F_\text{agent} \equiv 0$ 
 \blue{ /*update agent*/}}
                        \STATE Update $\pi(a|s)$ with $D_\text{model}$ and preferred algorithms (e.g. SAC approach~\citep{haarnoja2018soft} in RL) if needed 
                    \ENDIF
                \ENDFOR
            \end{algorithmic}
        \end{algorithm}

\section{Experiments}
\label{sec:experiments}
    \begin{table*}[ht]
    \vspace{15pt}
    	\caption{Tasks in RWRL benchmark and their modified physical parameters. The function $\texttt{linspace}(a, b, n)$ returns $n$ values evenly spaced over $[a, b]$.}
    	\begin{center}
    		\begin{tabular}{llllll}
    			\toprule
    			\textbf{Task Name} &  \textbf{dim($\mathcal{S}$)} & \textbf{dim($\mathcal{A}$)} & \textbf{Modified parameters} & \textbf{Training Values} & \textbf{Testing Values}  \\
    			 \midrule
    			\multirow{4}{*}{\shortstack{\texttt{walker\_stand}\\\texttt{walker\_walk}}} & \multirow{4}{*}{24}  & \multirow{4}{*}{6} & thigh length   & 0.225 & \texttt{linspace}(0.1, 0.7, 20) \\
    		    &&& torso length  & 0.3 & \texttt{linspace}(0.1, 0.7, 20) \\
    			&&& joint damping    & 0.1  & \texttt{linspace}(0.1, 10.0, 20)  \\
    			&&& contact friction  & 0.7  & \texttt{linspace}(0.01, 2.0, 20) \\
    			\midrule
    		   \multirow{3}{*}{\shortstack{\texttt{humanoid\_walk}}} & \multirow{3}{*}{45}  & \multirow{3}{*}{17}
                    & head size & 0.09 & \texttt{linspace}(0.01, 0.19, 20) \\
                    &&& contact friction & 1.0  & \texttt{linspace}(0.01, 2.5, 20) \\
                    &&& joint damping   & 1.0 & \texttt{linspace}(0.05, 1.2, 20) \\
    			\bottomrule
    		\end{tabular}
    	\end{center}
    	\label{tab:rwrl_env}
    \end{table*}

     \begin{figure}[ht!]
        \centering
        \includegraphics[width=0.5\textwidth]{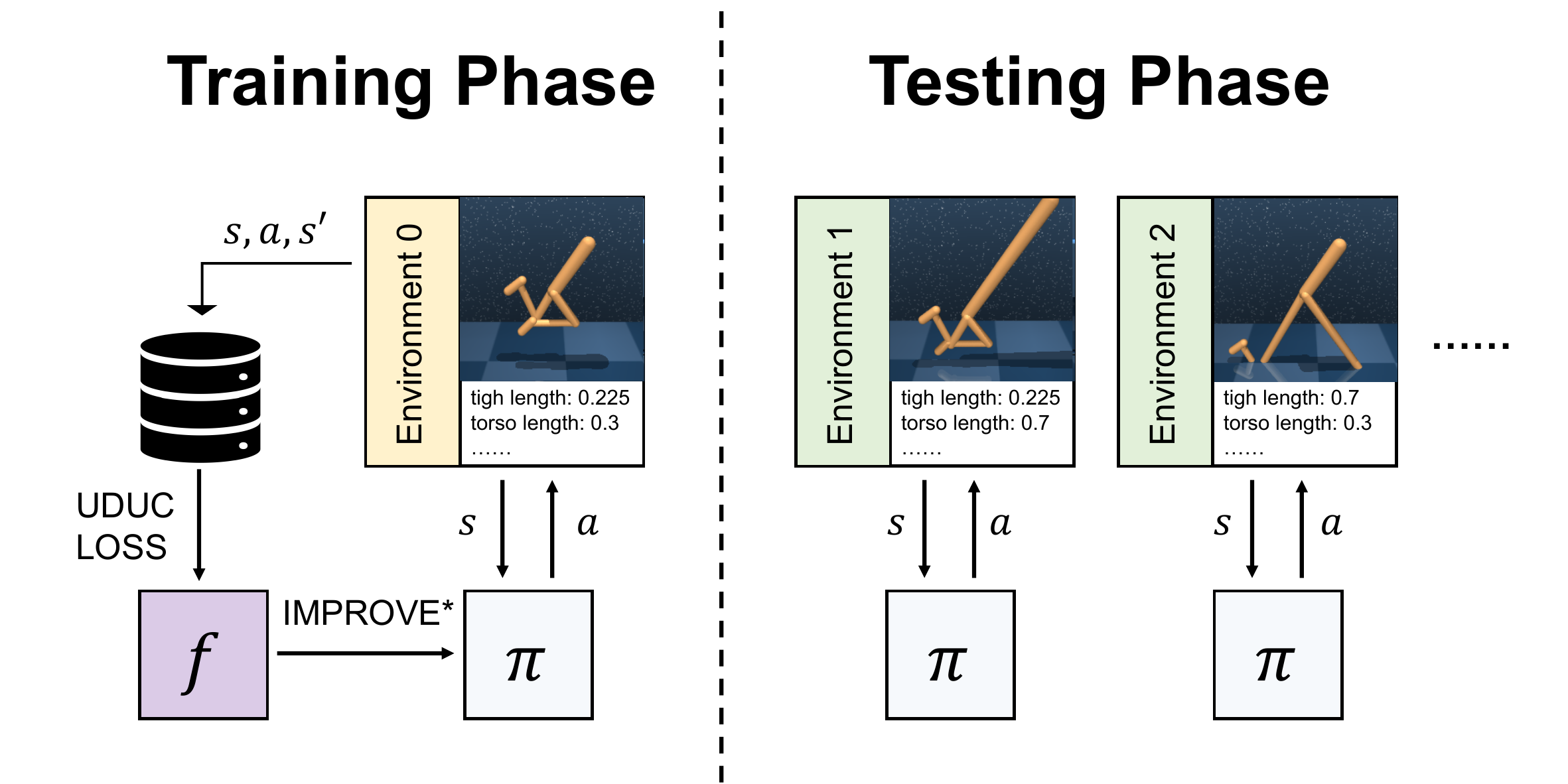}
        \caption{Visualization on the task setup. Notably, only one default environment is available during training, while the learned policy is evaluated on multiple varied testing environments. The approach to improve policy $\pi$ from learned transition function $f$ depends on specific algorithms, e.g. SAC~\citep{haarnoja2018soft} for reinforcement learning, CEM~\citep{rubinstein1999crossentropy} for model predictive control. }
        \vspace{15pt}
    	\label{fig:task_setup}
    \end{figure}

    \begin{figure}[ht!]
        \centering
        \begin{subfigure}[b]{1.0\linewidth}
            \centering        	    
            \includegraphics[width=\textwidth]{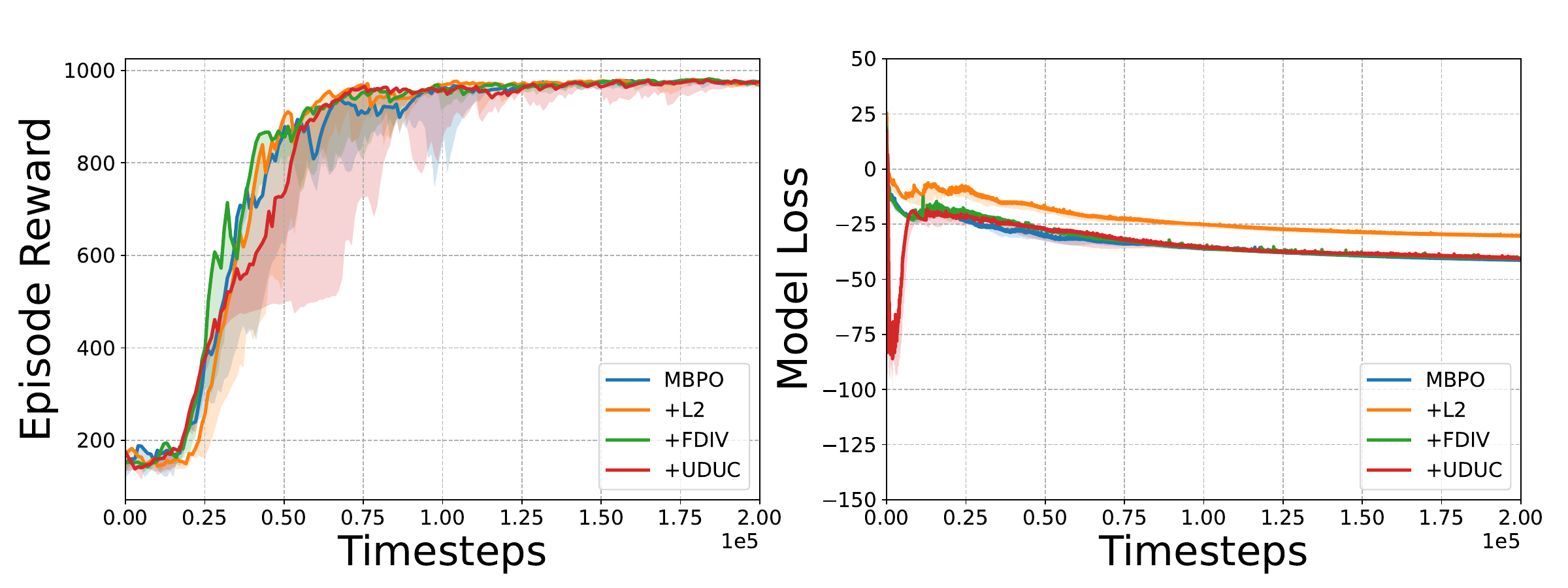}
    	    \caption{\texttt{walker\_stand}}
            \hfill
    	\label{fig:walker_stand_train}
        \end{subfigure}
        \begin{subfigure}[b]{1.0\linewidth}
            \centering        	    
            \includegraphics[width=\textwidth]{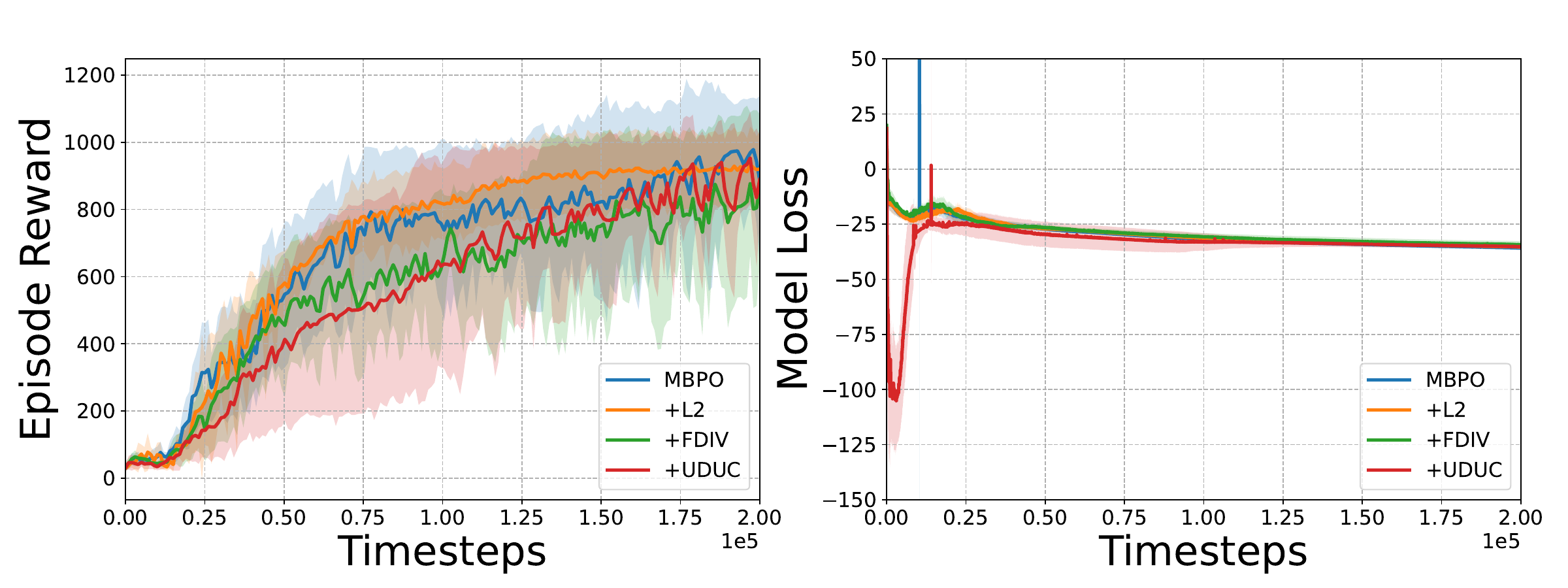}
    	    \caption{\texttt{walker\_walk}}
            \hfill
    	\label{fig:walker_walk_train}
        \end{subfigure}
        \begin{subfigure}[b]{1.0\linewidth}
    	\centering        	   
            \includegraphics[width=\textwidth]{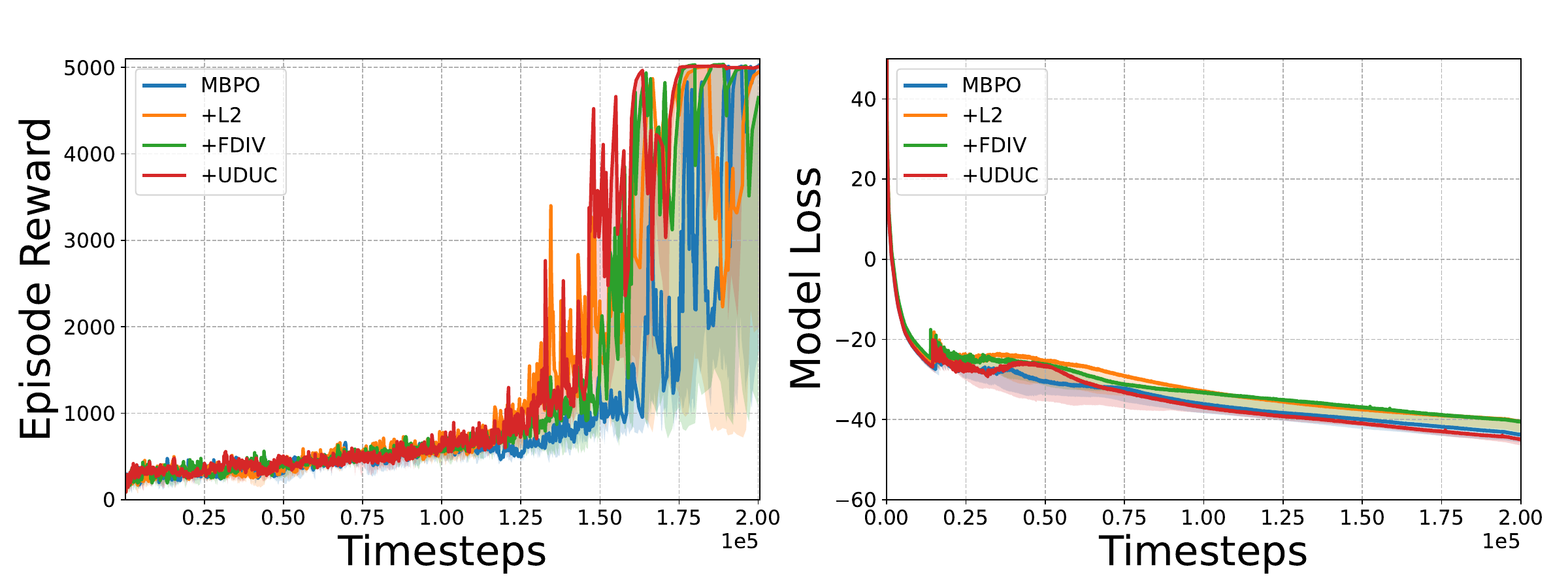}
    	    \caption{\texttt{humanoid\_walk}}
            \hfill
    	\label{fig:humanoid_walk_train}
        \end{subfigure}
    \caption{Training curves on robot control tasks. The x-axis is the environment time steps and the y-axis is the sum of rewards for each episode and the training loss for ensemble functions. All graphs are plotted with median and 25\%-75\% percentile shading across 5 seeds.}
    \vspace{15pt}
    \label{fig:rwrl_train}
    \end{figure}

    \begin{figure*}[ht!]
        \centering
        \begin{subfigure}[b]{1.0\linewidth}
            \centering        	    
            \includegraphics[width=\textwidth]{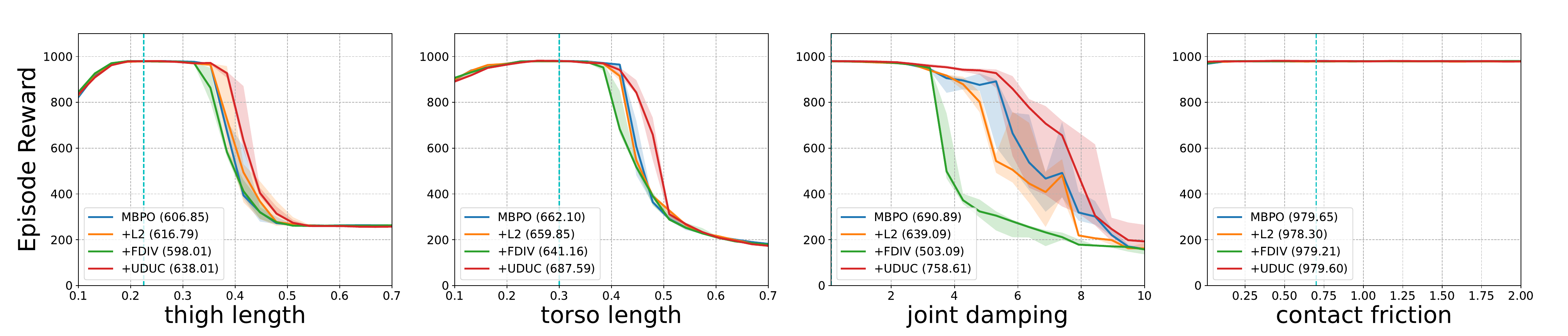}
    	    \caption{\texttt{walker\_stand}}
            \hfill
    	\label{fig:walker_stand_test}
        \end{subfigure}
        \begin{subfigure}[b]{1.0\linewidth}
            \centering        	    
            \includegraphics[width=\textwidth]{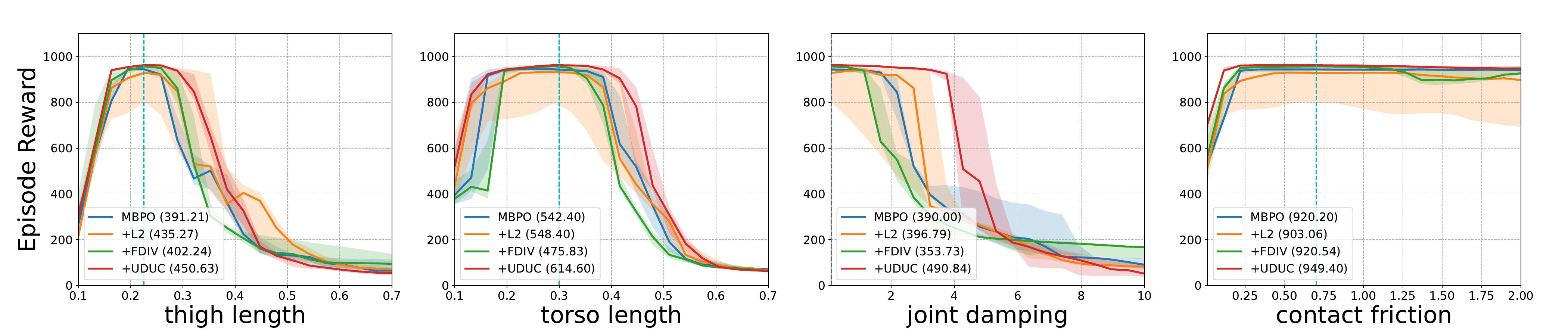}
    	    \caption{\texttt{walker\_walk}}
            \hfill
    	\label{fig:walker_walk_test}
        \end{subfigure}
        \begin{subfigure}[b]{1.0\linewidth}
    	\centering        	   
            \includegraphics[width=\textwidth]{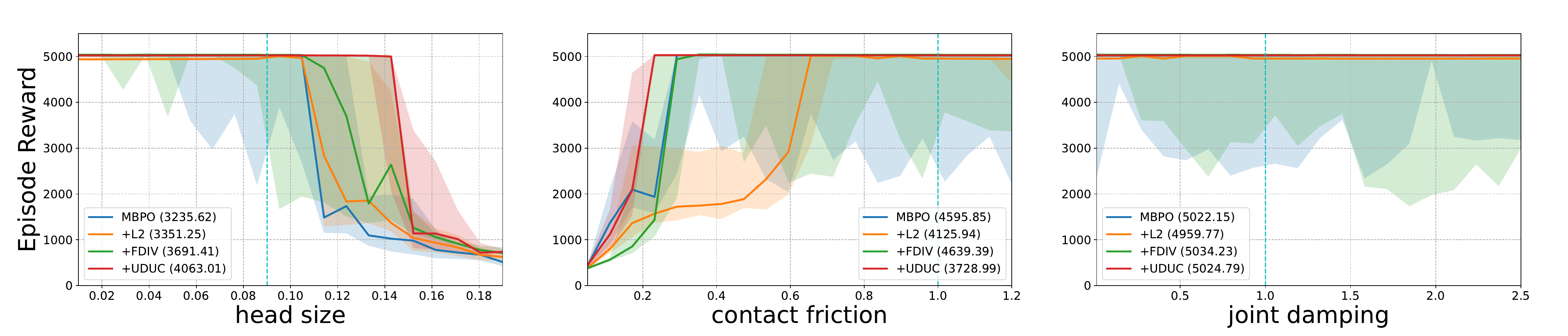}
    	    \caption{\texttt{humanoid\_walk}}
            \hfill
    	\label{fig:humanoid_walk_test}
        \end{subfigure}
    \caption{Testing curves on robot control tasks. The x-axis is the values of perturbed physical parameters and the y-axis is the episodic reward under the perturbed environments. The vertical dashed line denotes the nominal value during training. All graphs are plotted with the median and 25\%-75\%-quantile shading by running on 100 episodes. Robust-AUC is illustrated after each label in the graph.}
    \vspace{15pt}
    \label{fig:rwrl_test}
    \end{figure*}
    
    In this section, we evaluate the robustness of the UDUC-based methods on the Real-world Reinforcement Learning (RWRL) benchmark~\citep{dulac-arnold2020empirical}. Afterwards, we analyze the experimental results by visualizing the model parameters on the cart-pole~\citep{barto1983neuronlike} environment. 
     
\subsection{Experimental Setups}
\label{sec:experimental_setup}

    \paragraph{Task setup}
    RWRL is a continuous control benchmark for robotics, featuring multiple real-world challenges for RL algorithms, e.g. robustness, delays and safety. The physics simulator of RWRL is grounded on the MuJoCo engine~\citep{todorov2012mujoco}. 
    Upon this benchmark, we first train all model-based baselines on the nominal environment with default physical parameters, where the transition distribution is denoted as $\mathcal{T}_\text{train}(s'|s,a)$. Afterwards, some physical parameters are modified and therefore a set of testing environments is formalized. For each testing environment with the transition distribution $\mathcal{T}_\text{test}(s'|s,a)$, all trained baselines are tested on it without any fine-tuning, aiming to indicate the robustness against the mismatch in transition distributions. 

    In this paper, we conduct experiments on 2 robots (walker and humanoid) with 3 tasks: \texttt{walker\_stand}, \texttt{walker\_walk}, \texttt{humanoid\_walk}, with growing complexity in state and action space. The modified physical parameters and their values can be found in Table~\ref{tab:rwrl_env}. The range of the testing values significantly exceeds that of the training values, making the testing environments challenging. The details of the robots and tasks are introduced in Appendix~\ref{sec:introduction_on_robots_and_tasks}. We further exemplify the task setup in Figure~\ref{fig:task_setup}. During training, the transition $f$ learns the dynamics of the default walker robot and the policy $\pi$ learns to make the robot stand with the knowledge of $f$. During testing, the walker robot's physical parameters (e.g. thigh length, torso length) have been changed, and the learned policy $\pi$ is directly tested on those modified environments without tuning.

    \paragraph{Baselines} We consider model-based RL algorithms as baselines in this section. Viewing UDUC loss as an approach to regularize the model learning, we include other regularization approaches to compare their robustness. 
    
    \begin{itemize}
        \item MBPO: a model-based RL approach which combines probabilistic ensemble functions for transition modelling and SAC~\citep{haarnoja2018soft} for policy improvement. This baseline is adopted to indicate the performance without regularization techniques;
        \item MBPO + L2: an $L_2$-norm regularization is added in the model training process, expressed as $\mathcal{L}(\theta)=\sqrt{\theta^T \theta}$. $L_2$-norm regularization is proved to be one of the simplest and the most effective regularization methods in RL research~\citep{liu2021regularization};
        \item MBPO + FDIV: an objective to minimize the $f$-divergence~\citep{tiulpin2022greedy} is performed on learning ensemble functions. Notably, the work also aims to promote the diversity of ensemble functions but only regulates the mean predictions; 
        \item MBPO + UDUC: it follows Algorithm~\ref{alg:uduc_algorithm} to minimize UDUC loss when training models and adopts SAC to learn policy $\pi(a|s)$ at the same time. 
    \end{itemize}
    
    All baselines adopt the same neural network architectures as ensemble functions. The hyperparameters of all baselines are set equally for fair comparison, except for the regularization coefficients, which are tuned accordingly. The specific setups are clarified in Appendix~\ref{sec:hyperparameters_for_rwrl}.

\subsection{Main Results}
\label{sec:main_results}

    We first demonstrate the training performances in Figure~\ref{fig:rwrl_train}. Since all baselines are trained on the same default environment, they converge to almost the same episode reward in the end. Notably, at the early stage of training a walker robot, UDUC loss learns slower than baselines and indicates a larger uncertainty. At this stage, the diversity part of UDUC loss overwhelms the part of imitating the default environment, which is also shown in the curve of model loss that there is a dent at the beginning. For more complex robots like humanoids, regularization itself can benefit from reducing unnecessary information and speeding up the training process. A more detailed analysis of the training process is delivered in Appendix~\ref{sec:training_performances}. 

    Moving on to the testing phase, we organize evaluation results by different physical parameters as Figure~\ref{fig:rwrl_test}. We can first observe that the control performances become worse for all agents as the physical parameter deviates from the training value. Importantly, UDUC stays on top of other baselines for most of the time, indicating more robust control performance under various testing environments. Inspired by ~\citet{zhang2023robust}, we measure the robustness numerically with the metric Robust-AUC (noted in brackets in the legend), which is calculated as the relative area under the curve of the episode reward to the perturbed physical parameters. The Robust-AUC of UDUC clearly outperforms other baselines, especially for physical parameters that have large influences on the dynamics (e.g. joint damping in the walker robot and head size on the humanoid robot). 
    Notably, L2 and FDIV regularizations only improve the robustness in a few cases, but sometimes even impair the performance in comparison with the non-regularised agent on simple domains. 
     
     

\subsection{Ablation Study}
\label{sec:ablation_study}
    
    \begin{figure}[htb!]
        \centering
         \begin{subfigure}[b]{0.45\textwidth}
            \includegraphics[width=1.0\columnwidth]{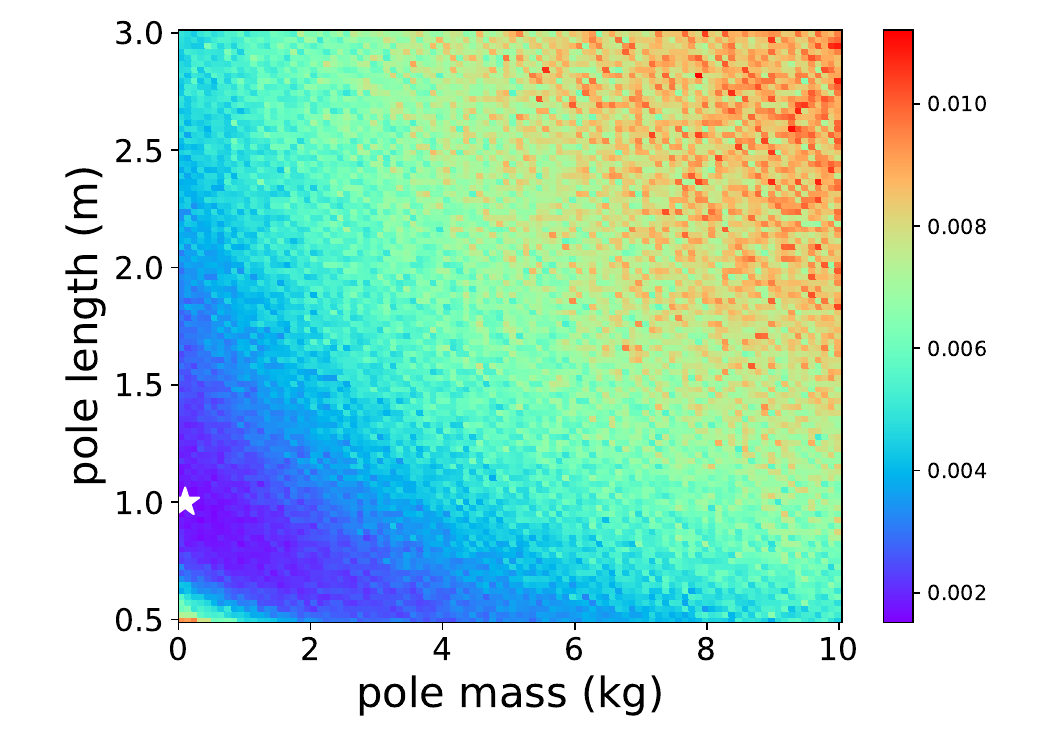}
            \caption{The average prediction error on states between training and testing physical parameters. The error is averaged over 100 random predictions. The star represents the training parameter.}
            \hfill
            \label{fig:cartpole_heatmap}
        \end{subfigure} \\
        \begin{subfigure}[b]{0.45\textwidth}
            \includegraphics[width=1.0\columnwidth]{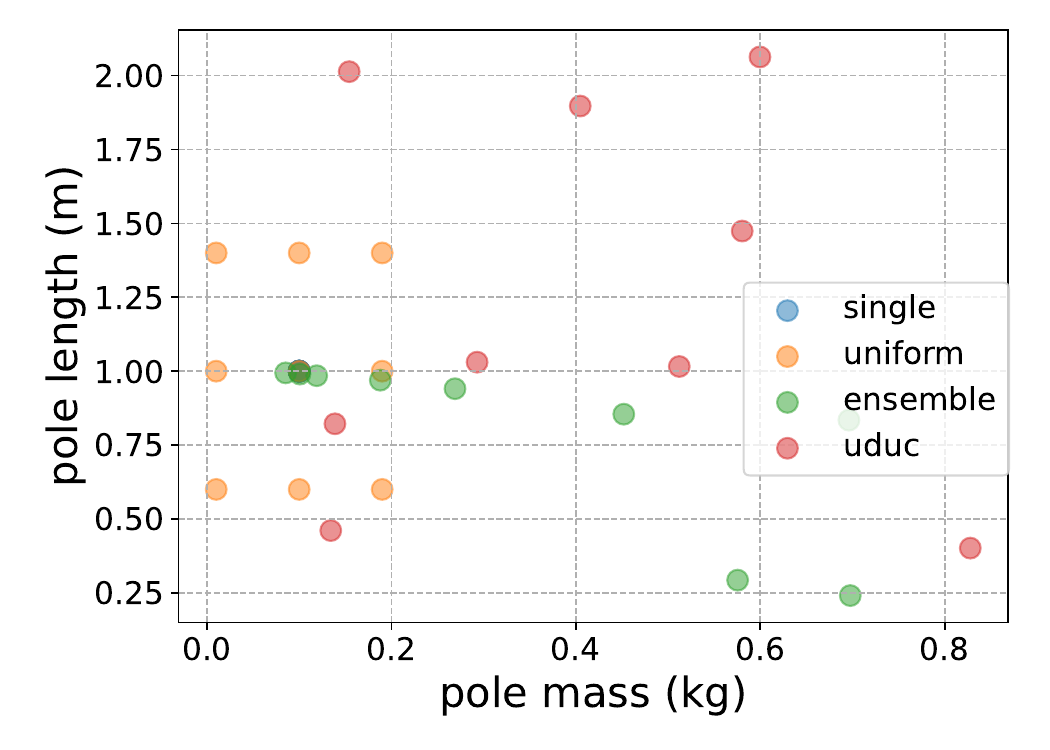}
            \caption{The learned parameters of the physical ensemble functions. There are 9 points for each method.}
            \hfill
            \label{fig:cartpole_weights9}
        \end{subfigure} \\
        \begin{subfigure}[b]{0.5\textwidth}
            \includegraphics[width=1.0\textwidth]{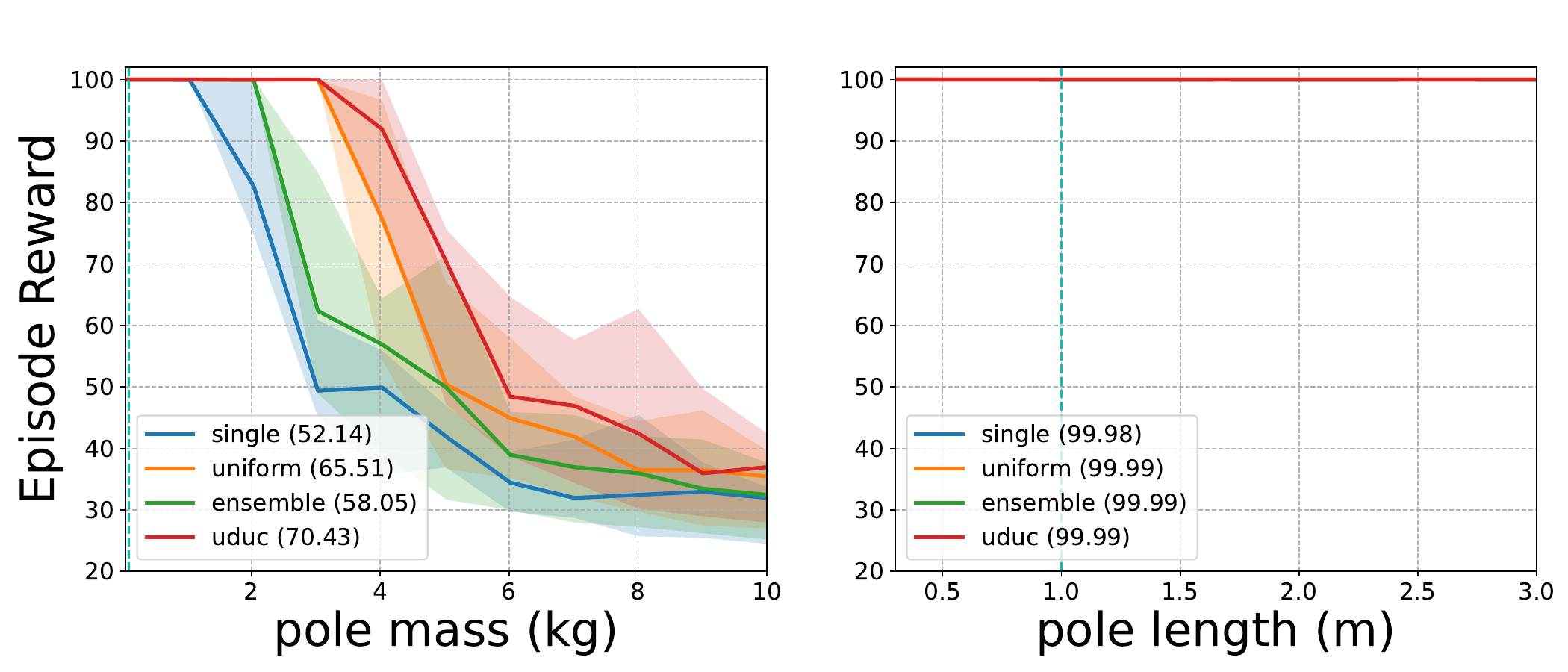}
            \caption{Testing curves on the cart-pole balance task, which has the same format as Figure~\ref{fig:rwrl_test}.}
            \hfill
    	   \label{fig:cartpole_test}
        \end{subfigure} \\
    \caption{Testing analysis on the cart-pole balance task.}
    \vspace{15pt}
    \label{fig:cartpole_results}
    \end{figure}

    \begin{figure*}[htb!]
        \centering
        \begin{subfigure}[b]{0.33\textwidth}
            \includegraphics[width=1.0\columnwidth]{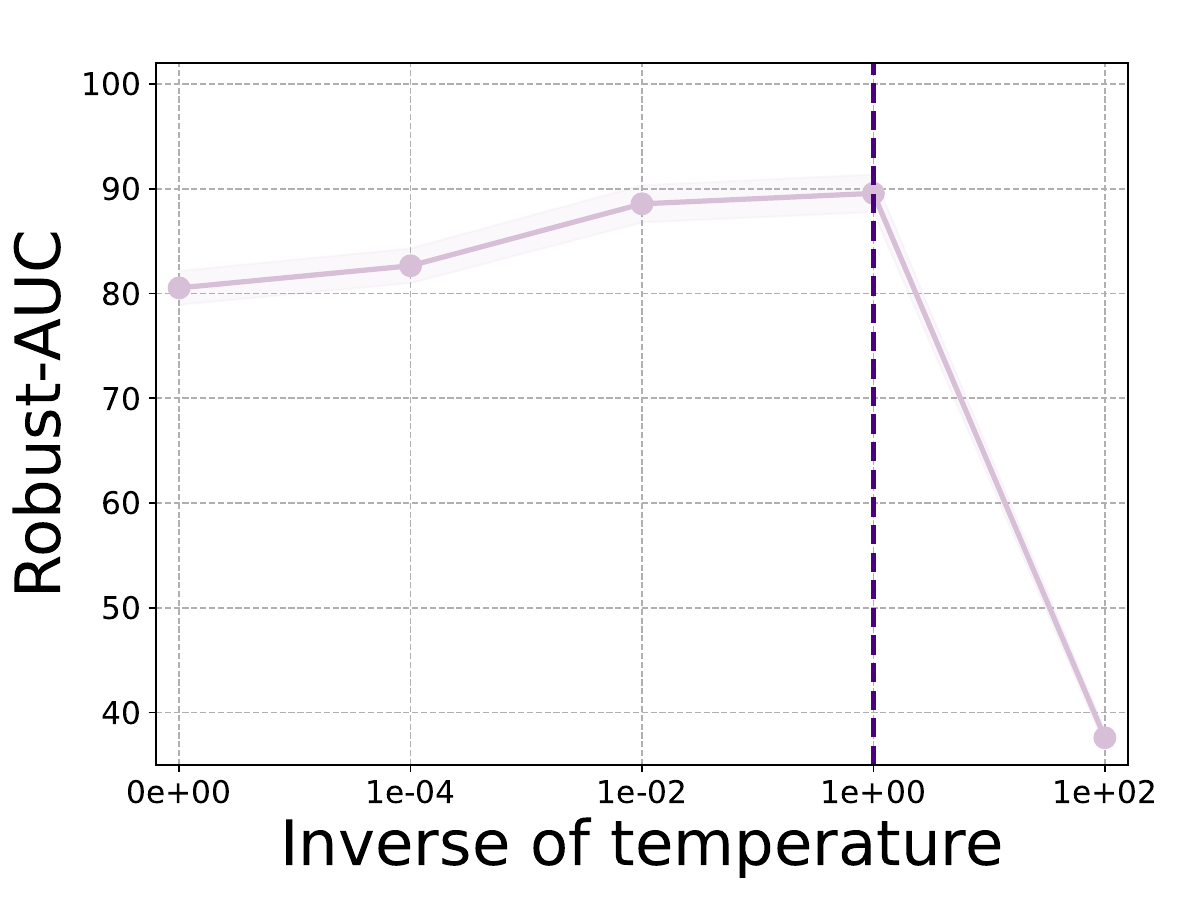}
            \label{fig:temperature} 
        \end{subfigure}
        \begin{subfigure}[b]{0.33\textwidth}
            \includegraphics[width=1.0\columnwidth]{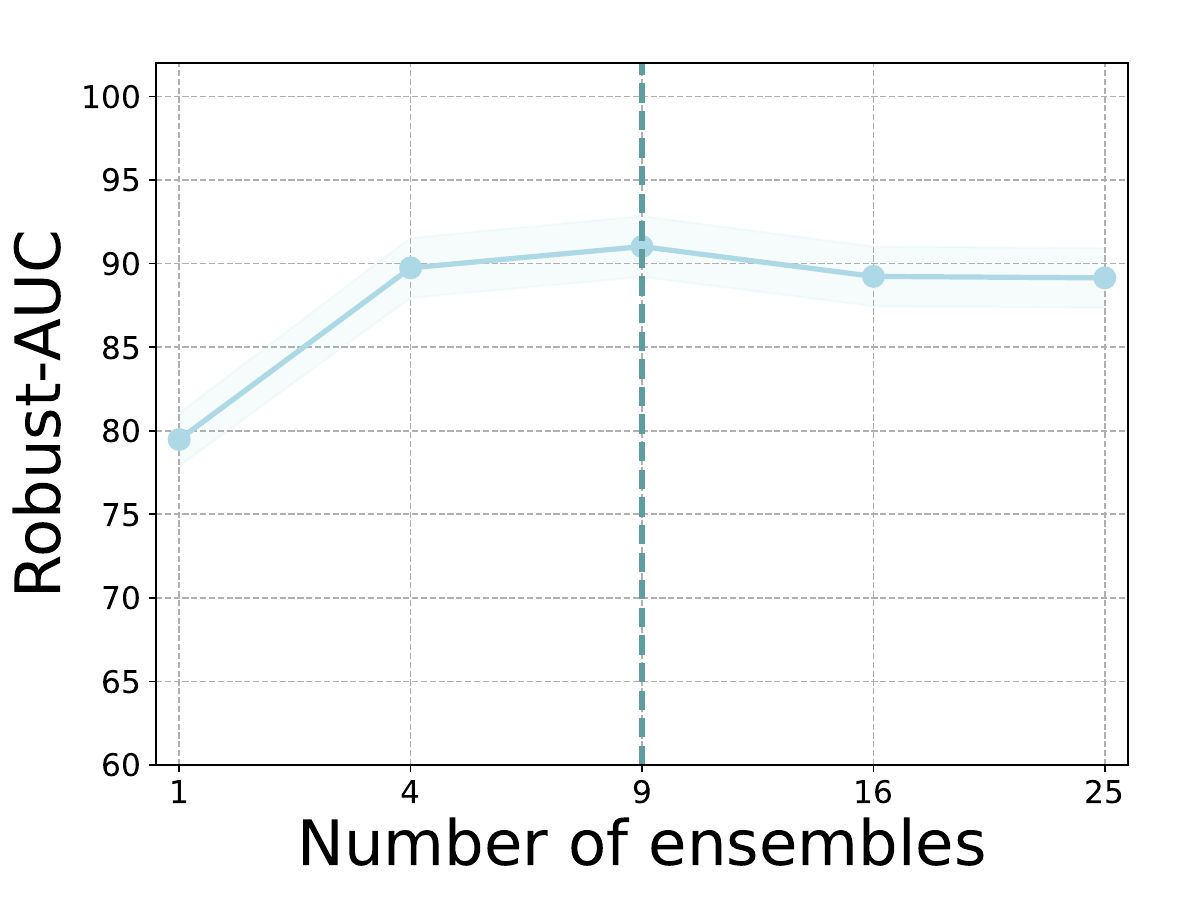}
            \label{fig:number} 
        \end{subfigure}
        \begin{subfigure}[b]{0.33\textwidth}
            \includegraphics[width=1.0\columnwidth]{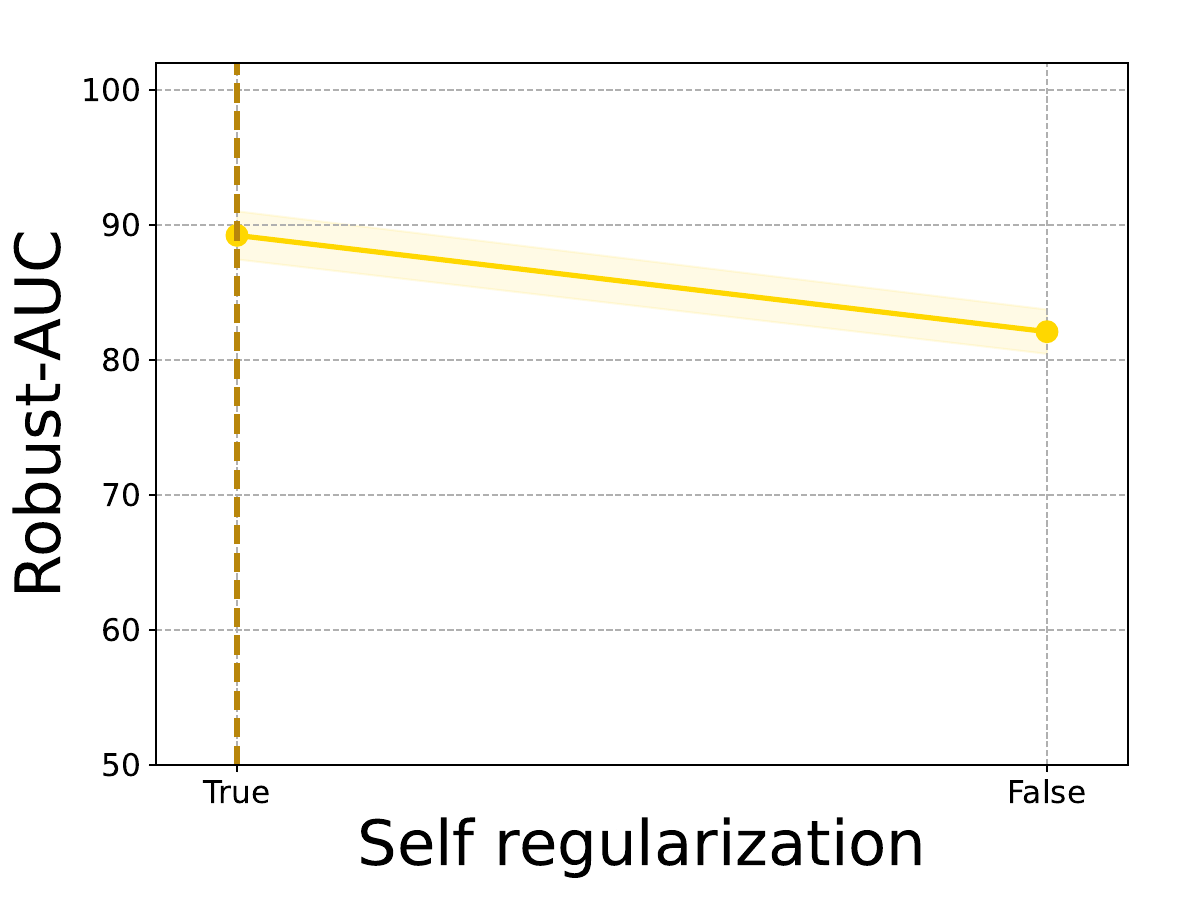}
            \label{fig:self_regularization} 
        \end{subfigure}
        \caption{Robust-AUC w.r.t. different designs of UDUC loss.}
        \vspace{15pt}
        \label{fig:ablation_study}
    \end{figure*}

    To fully understand why UDUC loss can provide additional robustness in complex continuous control tasks, we visualize the learning outcomes of UDUC loss on a toy example in this section. Specifically, we would like to answer the following questions: (1) what is the distribution of the learned model parameters? (2) how does the temperature $\tau$ act on the robustness? (3) does the self-regularization promote overall diversity? (4) do more ensemble members bring more robustness? 

    \paragraph{Task setup}
    For this toy example, we select the task of balancing a cart-pole with varying mass $m$ and length $l$~\citep{brockman2016openaia}. The environment has a 4-dim state space and a 1-dim action space. The reward is $1$ for an upright state otherwise 0. The maximum episode length is $100$. 
    Suppose the task holds a fixed noise, and its true dynamics with $f_t(s'|s,a;m,l)$ are determined by the mass and length, for simplicity of analysis, we directly assign the probabilistic ensemble model $b$ the same structure $f_b(s'|s,a;\theta_b)= f_t(s'|s,a;m_b,l_b)$ with parameters $\theta_b = (m_b, l_b)$ to learn. To avoid the influence of other learning processes, we adopt an MPC-based approach for control (see Appendix~\ref{sec:model_predictive_control} for details). The number of ensemble members $B=9$ and temperature $\tau=1$ unless otherwise stated. The environmental parameters are set to $m=0.1, l=1.0$ for training, and $m \in [0.05, 10.0],  l \in [0.3, 3.0]$ for testing.

    \textbf{Learned parameters}
    Since we adopt the physical functions as the ensemble functions, we can directly visualize the learned parameters in Figure~\ref{fig:cartpole_weights9}. The compared methods are (i) ensemble functions trained with negative log-likelihood only ("ensemble"); (ii) all 9 ensemble members are set to the same parameters as the training environment ("single") (iii) all 9 ensemble members are evenly distributed on the parameter space ("uniform"), centring on the training parameter.  
    It can be observed that UDUC loss learns more diverse parameters and occupies larger parameter space compared with other baselines. 
    Figure~\ref{fig:cartpole_heatmap} tells the prediction mismatch between training (marked in star) and testing environments for reference, which is calculated by $\mathbb{E}_{s,a}[(f_t(s,a; m_\text{train}, l_\text{train})-f_t(s, a; m_\text{test}, l_\text{test}))^2]$, where $s,a$ are randomly sampled from the state and action space. There is a larger mismatch in the top right corner, and only UDUC loss learns members in this region. Combined with the utilization of the trajectory sampling approach (introduced in Section~\ref{sec:practical_robust_control_approach}), there exists the potential for controllers to encounter transitions from this region, leading to robust performance under testing with unseen environmental parameters.
    
    We further plot the testing curves in Figure~\ref{fig:cartpole_test}. Both the curve and Robust-AUC value indicate that UDUC loss learns more robust ensemble parameters. More interestingly, the two physical parameters influence the robustness at different levels. UDUC pays more attention to pole mass with members $\sim 8$ times training values in the pole mass parameter, while $\sim 2$ times in the pole length parameter.


    \textbf{Temperature $\tau$}
        As explained in Section~\ref{sec:understanding_uduc_loss}, the temperature $\tau$ in UDUC loss is connected with the robust radius $\eta$ of the dynamics mismatch. 
        As the temperature decreases, the corresponding robust optimization problem exhibits an increase in robust radius, which also means the solutions could be more conservative. We calculate the Robust-AUC under different values of temperature $\tau$ and plot on the left of Figure~\ref{fig:ablation_study}. Notably, the x-axis is the inverse of temperature $\tau$, which is proportional to the robust radius $\eta$. There exists a sweet spot in the temperature, meaning it doesn't impact the control performance too much but could maintain a larger range of robustness.

    \textbf{Number of ensemble members $B$}
        We plot how the robustness is influenced by the number of ensemble functions in the middle of Figure~\ref{fig:ablation_study}. As the number of ensemble members increases, the robustness increases at the beginning and gradually saturates. Ideally, more ensemble members can occupy more parameter space, but it could burden the training of UDUC loss by contrasting in too many directions.

    \textbf{Regularization on self-prediction}
        As introduced in Section~\ref{sec:implementing_uduc_loss}, UDUC loss utilizes the technique of "self-regularization" in practice, which adds the own predictions from $f(s'|s,a;\bar{\theta}_b)$ to regulate the learning of model $b$. Most previous related work~\citep{wang2019nonlinear, tiulpin2022greedy} remove this term from regularizers for pure negative samples. However, \citet{wu2023understanding} demonstrates that InfoNCE loss can gain additional robustness from the self-regularization term under the appropriate temperature, which is also true for UDUC loss. 
        The right result of Figure~\ref{fig:ablation_study} shows that self-regularization does help the overall performance. It can be understood in a way that the member $i$ also contrasts its past prediction generated by its target network. This essentially provides a larger set of transitions, resulting in more robust controllers. 

        The key takeaways from the ablation study are as follows: self-regularization and the use of a large number of ensemble models are crucial for enhancing the robustness of the controller. Additionally, the temperature parameter needs to be carefully tuned, as discussed in Section~\ref{sec:hyperparameters_for_rwrl}.

\section{Related Work}
\label{sec:related_work}

    Robustness is a fundamental dimension to evaluate a control method since real-world systems are usually noisy and non-stationary. This topic is well-studied in both Optimal Control (OC)~\citep{ackermann1993robust} and Reinforcement Learning (RL)~\citep{iyengar2005robust, wiesemann2013robust, moos2022robust, zhang2023robust} fields. The overall vision is to derive a control policy $\pi^*(a|s)$ that can resist the perturbations in the transition function $\mathcal{T} \in \mathbb{T}$. In mathematics, it is equivalent to $\pi^* = \text{argmax}_{\pi} \min_{\mathcal{T} \in \mathbb{T}} \mathbb{E}_{ \pi, \mathcal{T} } \left[ \sum_{t=0}^{+\infty} \gamma^t r(s_t, a_t) \right]$. With the recent successes in the field of machine learning, it becomes practical and promising to learn transition functions $f(s'|s,a;\theta)$ from the data, and apply such functions to achieve the final control policy $\pi(a|s)$, denoted as Model-based RL (MBRL) and Learning-based Control (LBC) in both fields. As a result, the policy is strongly related to the parameters $\theta$ of the transition functions, raising a great concern to learn both accurate and robust transition functions, which directly motivates this research. 
 
    Regularization plays a key role in improving the robustness of learning-based methods. The Euclidean norm is one of the most widely applied regularization methods, which also dominates the learning of policy functions in the field of RL~\citep{liu2021regularization}. However, as shown in Section~\ref{sec:experiments}, it fails to maintain robustness on probabilistic ensemble transition functions, which is one of the state-of-the-art modelling approaches to capture both epistemic uncertainty and aleatoric uncertainty. 
    A line of work strives to promote diversity for ensemble learning~\citep{wang2019nonlinear, kariyappa2019improvinga, ross2020ensembles, rame2020dice, tiulpin2022greedy}. Unlike InfoNCE considering all ensemble members' predictions, most of these methods exclude the predictions of the trained member when calculating the diversity regularization. This could lead to a smaller robust radius under perturbed testing environments, which is analyzed in Section~\ref{sec:ablation_study}. We select a representative and concise work~\citet{tiulpin2022greedy} among them as baselines in the experiments. This work manages to increase the diversity of deep ensembles by the marginal gain on the negative $f$-divergence, without heavy adversarial dataset~\citep{ kariyappa2019improvinga, ross2020ensembles} or expensive generative models~\citep{rame2020dice}.

    Contrastive learning has been involved in the RL field recently. CURL~\citep{laskin2020curl} contrasts different images with InfoNCE loss~\citep{oord2018representation} and momentum target~\citep{he2020momentum} to learn a representative low-level space for further controls. The success of CURL is strongly related to better image embedding and less on the control rule itself. ~\citet{eysenbach2019diversity} maximizes the mutual information between states and skills by contrasting randomly sampled skills and encourages the agents to explore the uncovered skills, which mainly promotes the exploration but not robustness. 

\section{Conclusion}
\label{sec:conclusion}

    In conclusion, in this paper, we propose UDUC loss as an alternative objective to train probabilistic ensemble transition models inspired by InfoNCE loss. The diverse transition models can further enhance the robust control performance, which can be understood from domain randomization and robust optimization perspectives. Furthermore, we evaluate the robustness of UDUC on the challenging RWRL benchmark, where large environmental mismatches exist, and UDUC can consistently lead to more robust controllers. 

    For future work, we plan to validate UDUC with a more diverse set of low-level control algorithms, including the optimization-based optimal control and evolution strategy, to prove its wide applicability as L2 regularization. Besides, it will be interesting to combine UDUC with domain randomization methods, gathering both virtual and real multiple environmental interactions for a larger range of robustness.



\begin{ack}
This work is fully supported by the European Union’s Horizon 2020 research and innovation program under the Marie Skłodowska-Curie grant agreement No. 953348 (ELO-X).    
\end{ack}



\bibliography{ref}


\newpage
\onecolumn

\appendix

\begin{center}
  \textbf{\Large UDUC: An Uncertainty-driven Approach for Learning-based Robust Control\\(Supplementary Material)}
\end{center}

\section{Additional Algorithm Details}
\label{sec:additional_algorithm_details}

\subsection{Proof of Proposition~\ref{prop:understanding_uduc_loss}}
\label{sec:proof_of_proposition}

    In the case of UDUC loss, we first denote the training environment's transition distribution $\mathcal{T}_\text{train}(s'|s,a)$ ($\mathcal{T}_\text{train}$ in short), possible testing environment's transition distribution $\mathcal{T}_\text{test}(s'|s,a)$ ($\mathcal{T}_\text{test}$ in short) and probabilistic ensemble functions with target parameters $f(s'|s,a; \bar{\theta}) = \frac{1}{B} \sum_{b=1}^B f_b(s'|s,a; \bar{\theta}_b)$. We formally formulate a robust optimization problem and prove its equivalence to UDUC loss (Equation~\ref{eq:uduc_loss}) in the following Proposition.

    \setcounter{proposition}{0}
    \begin{proposition}[Understanding UDUC loss]
    \label{prop:understanding_uduc_loss_2}
        Minimizing UDUC loss is equivalent to a robust optimization problem:
        \begin{equation*}  
             \begin{aligned}
                 \min_{\theta_b} \mathcal{L}_{\text{RO}}(\theta_b,  s, a) &= \min_{\theta_b} (1 - \frac{1}{\tau}) \underbrace{\mathbb{E}_{s' \sim \mathcal{T}_\text{train}} [\mathcal{L}_{\text{PE}}(\theta_b, s, a, s')]}_{\text{negative log-likelihood term}} + \frac{1}{\tau} \underbrace{\big( \mathbb{E}_{s' \sim f(s'|s,a;\bar{\theta})} [-\mathcal{L}_{\text{PE}}(\theta_b, s, a, s')] -  \max_{\mathcal{T}_\text{test}} \mathbb{E}_{s' \sim \mathcal{T}_\text{test}} [-\mathcal{L}_{\text{PE}}(\theta_b, s, a, s')] \big)}_{\text{regularization term}}\\
                & \textit{s.t.} \quad D_{\text{KL}}(\mathcal{T}_\text{test}||\mathcal{T}_\text{train}) \le \eta, \eta \approx \mathbb{V}_{s'\sim f(s'|s,a)} [\mathcal{L}_{\text{PE}}(\theta_b, s, a, s')]/2 \tau^2,
             \end{aligned}
        \end{equation*} 
    \end{proposition}
    where $\mathcal{L}_{\text{RO}}(\theta_b,  s, a)$ is the target of this robust optimization problem, consisting of a negative log-likelihood term and a regularization term. 

    \begin{proof}
    
    Rethinking UDUC loss in Equation~\ref{eq:uduc_loss} $\mathcal{L}_{\text{UDUC}}(\theta_b, s, a, \mathcal{X}) = (1-\frac{1}{\tau})\mathcal{L}_{\text{PE}}(\theta_b, s, a, s_b') + \mathcal{L}_{\text{InfoNCE}}(\theta_b, s, a, \mathcal{X})$, we can first rewrite it into the expectation form as $\mathcal{L}_{\text{UDUC-e}}(\theta_b, s, a) = (1-\frac{1}{\tau})\mathcal{L}_1(\theta_b, s, a) + \mathcal{L}_2(\theta_b, s, a)$ where $\mathcal{L}_1(\theta_b, s, a) =  \mathbb{E}_{s' \sim \mathcal{T}_\text{train}} [\mathcal{L}_{\text{PE}}(\theta_b, s, a, s')] $ and $ \mathcal{L}_2(\theta_b, s, a) =  \mathbb{E}_{s^+ \sim \mathcal{T}_\text{train}, s^- \sim f(s'|s,a';\bar{\theta})}[\mathcal{L}_{\text{InfoNCE}}(\theta_b, s, a, \mathcal{X})]$. Notably, $\mathcal{L}_1(\theta_b, s, a)$ is the same as the negative log-likelihood term in $\mathcal{L}_\text{RO}(\theta_b, s, a)$. Therefore, we only need to prove that $\min \mathcal{L}_2(\theta_b, s, a)$ is equivalent to the reduced optimization problem as follows: 
    \begin{equation} 
    \label{eq:reduced_optimization_problem}
             \begin{aligned}
                 \min_{\theta_b} \mathcal{L}_{\text{RO-r}}(\theta_b,  s, a) &= \min_{\theta_b} \frac{1}{\tau} \underbrace{\big( \mathbb{E}_{s' \sim f(s'|s,a;\bar{\theta})} [-\mathcal{L}_{\text{PE}}(\theta_b, s, a, s')] -  \max_{\mathcal{T}_\text{test}} \mathbb{E}_{s' \sim \mathcal{T}_\text{test}} [-\mathcal{L}_{\text{PE}}(\theta_b, s, a, s')] \big)}_{\text{regularization term}}\\
                & \textit{s.t.} \quad D_{\text{KL}}(\mathcal{T}_\text{test}||\mathcal{T}_\text{train}) \le \eta, \eta \approx \mathbb{V}_{s'\sim f(s'|s,a)} [\mathcal{L}_{\text{PE}}(\theta_b, s, a, s')]/2 \tau^2,
             \end{aligned}
        \end{equation} 
         where $\mathcal{L}_{\text{RO-r}}(\theta_b,  s, a)$ only preserves the regularization term in $\mathcal{L}_{\text{RO}}(\theta_b,  s, a)$.

    We can further rewrite $\mathcal{L}_2(\theta_b, s, a)$ in the following way:
    \begin{equation*}  
             \begin{aligned}
                 \mathcal{L}_2(\theta_b, s, a) &=  \mathbb{E}_{s^+ \sim \mathcal{T}_\text{train}, s^- \sim f(s'|s,a';\bar{\theta})}[\mathcal{L}_{\text{InfoNCE}}(\theta_b, s, a, \mathcal{X})] \\
                 &= - \log \mathbb{E}_{s^+ \sim \mathcal{T}_\text{train}} [\exp(-\mathcal{L}_{\text{PE}}(\theta_b, s, a, s^+)/\tau)]  + \log \mathbb{E}_{s^- \sim f(s'|s,a';\bar{\theta})} [\exp(-\mathcal{L}_{\text{PE}}(\theta_b, s, a, s^-)/\tau)] \\
                 &= - \big[- \log \mathbb{E}_{s^+ \sim f(s'|s,a';\bar{\theta})} [\exp(-\mathcal{L}_{\text{PE}}(\theta_b, s, a, s^+)/\tau)] + \log \mathbb{E}_{s^- \sim \mathcal{T}_\text{train}} [\exp(-\mathcal{L}_{\text{PE}}(\theta_b, s, a, s^-)/\tau)]\big] \\
                 &= - \mathcal{L}_\text{InfoNCE-alt} (\theta_b, s, a).
             \end{aligned}
        \end{equation*} 
   Therefore, minimizing $\mathcal{L}_2(\theta_b, s, a)$  is equivalent to maximizing another InfoNCE-styled loss $\mathcal{L}_\text{InfoNCE-alt} (\theta_b, s, a)$, where positive samples are drawn from $f(s'|s,a';\bar{\theta})$ and negatives from $\mathcal{T}_\text{train}$. 
    The following proof is built on the previous work~\citep{wu2023understanding} that has shown the equivalence between InfoNCE loss and robust the outcome of an optimization as follows:
    \begin{theorem}[CL-RO Equivalance~\citep{wu2023understanding}]
    \label{thereom: cl-ro_equivalance}
        For the expected version of InfoNCE loss $\mathcal{L}_{\text{InfoNCE-e}}(\theta, c) = - \log \frac{ \mathbb{E}_{x^+ \sim p(x^+|c)} [\exp(g_{\theta}(x, c)/\tau)] }{\mathbb{E}_{x^- \sim q(x)} [\exp(g_{\theta}(x^-, c)/\tau)]}$, it is equivalent to the outcome of an optimization problem:
        
        \begin{equation*}  
             \begin{aligned}
                \mathcal{L}_{\text{opt}}(\theta,  c) = \max_z  - \mathbb{E}_{x \sim p(x|c)} [g_{\theta}(x, c)] /\tau +  \mathbb{E}_{x \sim z(x))}[g_{\theta}(x, c)]  /\tau,  
 \quad \textit{s.t.} D_{\text{KL}}(z||q) \le \eta,
             \end{aligned}
        \end{equation*} 
        
        with the relation as $\mathcal{L}_{\text{opt}}(\theta,  c) = \mathcal{L}_{\text{InfoNCE-e}}(\theta, c) + \textit{const}, 
                 \tau \approx \sqrt{\mathbb{V}_{x\sim q(x)} [g_{\theta}(x, c)]/2\eta}.$
    \end{theorem}
    
    In the case of the MDP formulation, the context $c$ is actually the state and action pair $(s,a)$, the next state $s'$ as the sample $x$ and the representation function $g_{\theta_b}(x, c) =-\mathcal{L}_{\text{PE}}(\theta_b, s, a, s')$. The positive distribution $p(x|c)$ is equivalent to probabilistic ensemble functions with target parameters $f(s'|s,a; \bar{\theta}) = \frac{1}{B} \sum_{b=1}^B f_b(s'|s,a; \bar{\theta}_b)$ and the negative distribution $q(x)$ comes from the training transition distribution $\mathcal{T}_\text{train}(s'|s,a)$.
    According to the Theorem~\ref{thereom: cl-ro_equivalance}, the alternative InfoNCE loss $\mathcal{L}_\text{InfoNCE-alt} (\theta_b, s, a)$ can be transformed into the outcome of an optimization problem such that: 

    \begin{equation*}  
        \label{eq:ro}
             \begin{aligned}
                 \mathcal{L}_{\text{opt}}(\theta_b,  s, a) = - \frac{1}{\tau}\mathbb{E}_{s' \sim f(s'|s,a; \bar{\theta})} [-\mathcal{L}_{\text{PE}}(\theta_b, s, a, s')]  + \max_{\mathcal{T}_\text{test}} \frac{1}{\tau} \mathbb{E}_{s' \sim \mathcal{T}_\text{test}}[-\mathcal{L}_{\text{PE}}(\theta_b, s, a, s')] 
 \quad \textit{s.t.} D_{\text{KL}}(\mathcal{T}_\text{test}||\mathcal{T}_\text{train}) \le \eta.
             \end{aligned}
        \end{equation*} 

    Consequently, minimizing $ \mathcal{L}_2(\theta_b, s, a)$ is equivalent to maximizing $ \mathcal{L}_\text{opt}(\theta_b, s, a)$. This can be further transformed into minimizing $-\mathcal{L}_\text{opt}(\theta_b, s, a)$, which is exactly the case of $\min_{\theta_b}  \mathcal{L}_\text{RO-r}(\theta_b, s, a)$. 
    
    \end{proof}

\subsection{Control with Probabilistic Ensemble Functions}
\label{sec:control_with_probabilistic_ensemble_functions}

    \subsubsection{Model Predictive Control}
    \label{sec:model_predictive_control}

    Given a transition function $f(s'|s, a;\theta)$, deriving the policy $\pi(a|s)$ in the MPC is an optimization problem. To better handle the computational efficiency, cross-entropy-based methods~\citep{rubinstein1999crossentropy} are often considered. We detail the procedures in Algorithm~\ref{alg:mpc_pe} to generate controls with the learned ensemble function. In basic, the learned transition functions are used to generate trajectories for optimization. For each step, one ensemble member is randomly sampled, which ensures the multi-mode prediction in the next states and promotes robustness. The other propagation method is introduced in ~\citet{chua2018deep}.

    \begin{algorithm*}[ht!]
        \caption{Cross-Entropy with Probabilistic Ensembles}
            \begin{algorithmic}[1]
                \STATE \textbf{Input:} Ensemble transition function $f(s'|s, a;\theta)$ with parameters $\theta=\{\theta_1, \cdots, \theta_B\}$, policy function $\pi(a|s)$, environment dataset $\mathcal{D}_\text{env}=\{\}$, model dataset $\mathcal{D}_\text{model}=\{\}$, target update rate $\rho$, initial state $s$, model-agent update frequency $F_\text{model}$, $F_\text{agent}$, population size $N$, elite size $K$, horizon $H$, action dimension $A$
                \STATE Set target transition function's parameters $\bar{\theta} \leftarrow \theta$ 
                \STATE Set initial cross-entropy parameters $\mu \leftarrow [0,\dotsc,0]^T: \mathbb{R}^{H*A}$ the mean and $\Sigma \leftarrow I : (\mathbb{R}^{H*A})^2$ the variance
                \FOR{steps $i = 1,2,...$}
                    \FOR{iterations $j = 1,2,...$ \blue{ /*run cross-entropy*/}}
                        \STATE Generate $N$ samples $X_{1:N} \sim \mathcal{N}(\mu, \Sigma)$
                        \FOR{sample $n = 1,2,...N$}
                            \STATE Rollout a trajectory $(s_1, a_1, \cdots, s_H, a_H)$, where $a_t$ is directly accessed from $X_n[(t-1)*H:t*H]$, and $s_{t+1}$ sampled from $f_b(s'|s,a;\theta_b)$ where $b$ is a random select ensemble member. 
                            \STATE Calculate score $S_n \leftarrow \left[\exp \left(-{\sum_{t=1}^H {r(s_t, a_t)}}\right)\right]$
                        \ENDFOR
                        \STATE $X_{1:N} \leftarrow \textrm{sort}(X_{1:N})$ depending on $S_{1:N}$
                        \STATE $\mu \leftarrow \textrm{mean}\left(X_{1:K}\right)$
                        \STATE $\Sigma \leftarrow \textrm{var} \left(X_{1:K} \right)$
                    \ENDFOR
                    \STATE Execute $a = \mu$  in the training environment 
                    \STATE Observe reward $r$ and next state $s'$ \\
                    \STATE $\mathcal{D}_\text{env} \leftarrow \mathcal{D}_\text{env}  \cup \{ s, a, r, s'\} $ and $s \leftarrow s'$
                    \IF{$i \mod F_\text{model} \equiv 0$ \blue{ /*update model*/}}  
                        \STATE Update parameters $\theta$ by minimizing Equation~\ref{eq:uduc_loss}
                        \STATE Update target parameters $\bar{\theta} \leftarrow (1-\rho)\bar{\theta} + \rho \theta $ 
                        \STATE Rollout transitions $(s_n, a_n, r_n, s_{n+1})_{n=1}^N$ with $f(s'|s,a; \bar{\theta})$, $\pi(a|s)$ and  $r(s, a)$
                        \STATE Store in the model dataset $\mathcal{D}_\text{model} \leftarrow \mathcal{D}_\text{model} \cup \{ (s_n, a_n, r_n, s_{n+1})_{n=1}^N\}$
                    \ENDIF
                \ENDFOR
            \end{algorithmic}
        \label{alg:mpc_pe}
        \end{algorithm*}

    \subsubsection{Reinforcement Learning}
    \label{sec:reinforment_learning}

    For RL, the policy $\pi(a|s)$ is usually modelled as a neural network. The policy can be improved with actor-critic frameworks. Soft Actor Critic (SAC)~\citep{haarnoja2018soft} is one of the state-of-the-art algorithms in the field. We detail the procedures in Algorithm~\ref{alg:sac_pe} to learn $\pi(a|s)$ with the learned ensemble function. In basic, the learned transition functions are used to generate transitions $(s,a, s')$ and further stored in the dataset $\mathcal{D}_\text{model}$. SAC uses this data to update critic and actor parameters. The specific loss function $J_Q(\theta)$ $J_\pi(\phi)$ can be referred in ~\citet{haarnoja2018soft}.
    
    \begin{algorithm*}[ht!]
        \caption{Soft Actor Critic with Probabilistic Ensembles}
            \begin{algorithmic}[1]
                \STATE \textbf{Input:} Double action value $Q(s, a; \theta)$' with parameters $\theta^Q_1$, $\theta^Q_2$ ensemble transition function $f(s'|s, a;\theta)$ with parameters $\theta=\{\theta_1, \cdots, \theta_B\}$, policy function $\pi(a|s)$, environment dataset $\mathcal{D}_\text{env}=\{\}$, model dataset $\mathcal{D}_\text{model}=\{\}$, target update rate $\rho$, initial state $s$, model-agent update frequency $F_\text{model}$ and $F_\text{agent}$, SAC's learning rate for critic and actor network $\lambda_{Q}$ $\lambda_{\pi}$, and target network update ratio $\tau$
                \STATE Set target transition function's parameters $\bar{\theta} \leftarrow \theta$ 
                \FOR{steps $i = 1,2,...$}
                    \STATE Execute $a \sim \pi(a|s)$  in the training environment 
                    \STATE Observe reward $r$ and next state $s'$ \\
                    \STATE $\mathcal{D}_\text{env} \leftarrow \mathcal{D}_\text{env}  \cup \{ s, a, r, s'\} $ and $s \leftarrow s'$
                    \IF{$i \mod F_\text{model} \equiv 0$ \blue{ /*update model*/}}  
                        \STATE Update parameters $\theta$ by minimizing Equation~\ref{eq:uduc_loss}
                        \STATE Update target parameters $\bar{\theta} \leftarrow (1-\rho)\bar{\theta} + \rho \theta $ 
                        \STATE Rollout transitions $(s_n, a_n, r_n, s_{n+1})_{n=1}^N$ with $f(s'|s,a; \bar{\theta})$, $\pi(a|s)$ and  $r(s, a)$
                        \STATE Store in the model dataset $\mathcal{D}_\text{model} \leftarrow \mathcal{D}_\text{model} \cup \{ (s_n, a_n, r_n, s_{n+1})_{n=1}^N\}$
                    \ENDIF
                    \IF{$i \mod F_\text{agent} \equiv 0$ 
    \blue{ /*update agent*/}}
                        \STATE $\theta_i \leftarrow \theta_i - \lambda_Q \hat{\nabla}_{\theta_i} J_Q(\theta_i) \textrm{ for } i \in \{1, 2\}$
                        \STATE $\psi \leftarrow \psi - \lambda_\pi \hat{\nabla}_\psi J_\pi (\psi)$
                        \STATE $\bar{\theta}_i \leftarrow \tau \theta_i + (1 - \tau) \bar{\theta}_i \textrm{ for } i \in \{1, 2\}$
                    \ENDIF
                \ENDFOR
            \end{algorithmic}
        \label{alg:sac_pe}
        \end{algorithm*}    

\section{Additional Experimental Setups}
\label{sec:additional_experimental_setups}

\subsection{Introduction on Robots and Tasks}
\label{sec:introduction_on_robots_and_tasks}

\begin{figure}
    \centering
    \begin{subfigure}[b]{0.3\textwidth}
            \includegraphics[width=1.0\textwidth]{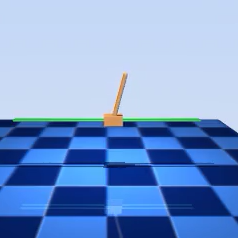}
            \caption{Cart-pole}
            \hfill
    	   \label{fig:cartpole}
        \end{subfigure} 
    \begin{subfigure}[b]{0.3\textwidth}
            \includegraphics[width=1.0\textwidth]{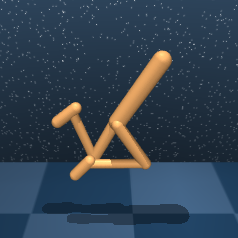}
            \caption{Walker}
            \hfill
    	   \label{fig:walker}
        \end{subfigure} 
        \begin{subfigure}[b]{0.3\textwidth}
            \includegraphics[width=1.0\textwidth]{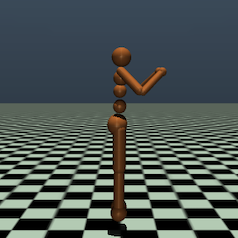}
            \caption{Humanoid}
            \hfill
    	   \label{fig:humanoid} 
        \end{subfigure} 
    \caption{Visualization of $3$ robots used in the experiments.} 
    \vspace{15pt}
    \label{fig:robots}
\end{figure}

    In this paper, the experiments mainly touch on $3$ different robots, the cart-pole, the walker and the humanoid. These $3$ robots are visualized in Figure~\ref{fig:robots}. The introduction of these robots is as follows:
    
    \begin{itemize}
        \item \textbf{Cart-pole} has an un-actuated pole on top of a moving cart. The \texttt{cartpole\_balance} task aims to balance the pole on the origin position with a horizontal force on the cart.  
        \item \textbf{Walker} is a planar walker to be controlled in 6 dimensions. The \texttt{walker\_stand} task requires an upright torso and some minimal torso height. The \texttt{walker\_walk} task encourages a forward velocity.
        \item \textbf{Humanoid} is a generic humanoid robot with a more complex state and action space than the cart-pole and the walker. The \textit{humanoid\_walk} tasks encourage a level of forward speed.
    \end{itemize}

    The specific reward function can be found in the original work of RWRL~\citep{dulac-arnold2020empirical}. Here, we set the maximum episode length as $1000$ for the walker and humanoid, and $100$ for the cart-pole. Therefore, the maximum possible episode reward is $1000, 5000, 100$ respectively. 

\subsection{Hyperparameters for RWRL}
\label{sec:hyperparameters_for_rwrl}

    For RWRL tasks, we learn a probabilistic ensemble model and use the model to train a neural network agent with the SAC algorithm in RL. The model has the structure of a feed-forward neural network to predict the mean and variance of the delta states. For the SAC agent, we adopt the double critic technique and with the agent generating a Gaussian control. 
    To compare all algorithms fairly, we set the model structures and hyperparameters equally, except for the regularization coefficients for different algorithms. All algorithms are trained with Adam optimizer \citep{kingma2015adam}. The full hyperparameters are shown in Table~\ref{tab:hyperparameters_rwrl}. For regularization coefficients, we manually tune them for each algorithm by increasing their values until the control performance on the nominal environment drops, and the values range in $\{0, 1e-6, 1e-5, 1e-4, 1e-3, 1e-2, 1e-1, 1e0\}$. The specific values can be found in Table~\ref{tab:reg_coef}. In the future, it can be automatically tuned by learning a non-regularized control policy and comparing the control performance between these agents. All experiments are carried out on NVIDIA GeForce RTX 2080 Ti and Pytorch 1.10.1. 
    
    \begin{table*}[ht!]
    	\caption{Hyperparameters for RWRL tasks.}
    	\begin{center}
    	\scalebox{1.0}{
			\begin{tabular}{lccc}
			\toprule
				\textbf{Hyperparameters} & \texttt{walker\_stand} & \texttt{walker\_walk} & \texttt{humanoid\_walk}  \\
			\midrule
                max training steps & 2e5 & 2e5 & 2e5 \\
                episode length & 1e3 & 1e3 & 1e3 \\
                dynamics network &   4x MLP(200) &  4x MLP(200) &  4x MLP(200) \\
                number of ensembles & 5 & 5 & 7 \\
                propagation method & random & random & random \\
                activation function & SiLU & SiLU & SiLU \\
                model learning rate & 3e-4 & 3e-4 &  3e-4 \\
                model batch size & 256 & 256 & 256 \\
                model update frequency & 250 & 250 & 250 \\
                effective model rollouts per step & 400 & 400 & 400 \\
                rollout schedule & [20,150,1,1] & [20,150,1,1] & [20,300,1,25] \\
                \blue{ /*for SAC agent*/} & \\
                critic network & 2 x MLP(1024)  & 2x MLP(1024)  & 2x MLP(1024) \\
                actor network & 2x 1024 & 2x 1024  & 2x 1024  \\
                discount factor $\gamma$ & 0.99 & 0.99 & 0.99 \\
                init temperature & 1.0 & 1.0 & 1.0 \\
                temperature learning rate & 3e-3 & 3e-3 & 1e-4  \\
                actor learning rate & 3e-4 & 3e-4 & 1e-4 \\
                actor update frequency & 4 & 4 & 1 \\
                critic learning rate & 3e-4 & 3e-4 & 1e-4 \\
                critic update frequency & 1 & 1 & 1 \\
                critic target update rate & 0.005 & 0.005 & 0.005 \\
                critic target update frequency & 4 &  4&  4 \\
                sac updates per step & 20 & 20 & 20 \\
                epochs to retain sac buffer & 1 & 1 & 5 \\
    			sac batch size & 256 & 256 & 256 \\
    			target entropy & -3 & -3 & -1 \\
			\bottomrule 
			\end{tabular}
			}
    	\end{center}
    	\label{tab:hyperparameters_rwrl}
    \end{table*}
    
    \begin{table*}[ht]
    	\caption{Regularization coefficients for all baseline methods.}
    	\begin{center}
    	 \scalebox{1.0}{
    		\begin{tabular}{lcccc}
    			\toprule
    			\multirow{2}{*}{\textbf{Task Name}} & 
    			\multicolumn{4}{c}{\textbf{Algorithms}} \\
    			& MBPO & +L2 & +FDIV &  + UDUC \\
    			\midrule	
    			\texttt{walker\_stand} & - & 1e-1 & 1e-2 & 1e-2  \\
    			\texttt{walker\_walk}  & - & 1e-2 & 1e-3 & 1e-3 \\
    			\texttt{humanoid\_walk} & - & 1e-2 & 1e-4 & 1e-3 \\
    			\bottomrule
    		\end{tabular}
    		}
    	\end{center}
    \label{tab:reg_coef}
    \end{table*}

\subsection{Hyperparameters for Cart-pole}
\label{sec:hyperparameters_for_cart-pole}

    For the cart-pole balance task, instead, we learn a physical-based ensemble model with parameters pole mass and pole length and use the model to drive a CEM controller in the MPC style. For the model, it has the structure of a feed-forward neural network to predict the mean of the next states. The full hyperparameters are shown in Table~\ref{tab:hyperparameters_cartpole}. 
    
    \begin{table*}[ht!]
    	\caption{Hyperparameters for cart-pole balance task.}
    	\begin{center}
    	\scalebox{1.0}{
			\begin{tabular}{lc}
			\toprule
			 \textbf{Hyperparameters} & \texttt{cartpole\_balance} \\  
			\midrule
                max training steps & 500 \\
                episode length & 100 \\
                number of ensembles & 9 \\
                propagation method & random \\
                \blue{ /*for CEM agent*/} & \\
                model learning rate & 1e-3 \\
                model batch size & 32 \\
                model update frequency & 100 \\
                planning horizon & 15 \\
                replan frequency & 1 \\
                optimization iterations & 5 \\
                elite ratio & 0.1 \\
                population size & 500 \\
                particles & 20 \\
			\bottomrule 
			\end{tabular}
			}
    	\end{center}
    	\label{tab:hyperparameters_cartpole}
    \end{table*}

\section{Additional Experimental Results}
\label{sec:additional_experimental_results}

\subsection{Training Performances}
\label{sec:training_performances}

     We provide additional training results regarding the neg log-likelihood, regularization, gradients norm and model training dataset. From the neg log-likelihood, we can see UDUC loss learns slower on the nominal model compared with other baselines, proving the contrastive loss is working at the beginning. This phenomenon is more obvious from the regularization and grad norm figure of the walker robot. For the figure of the size of the training dataset is to show that all models are trained on the same amount of data.

    \begin{figure*}[ht!]
        \centering
        \begin{subfigure}[b]{1.0\linewidth}
            \centering        	    
            \includegraphics[width=\textwidth]{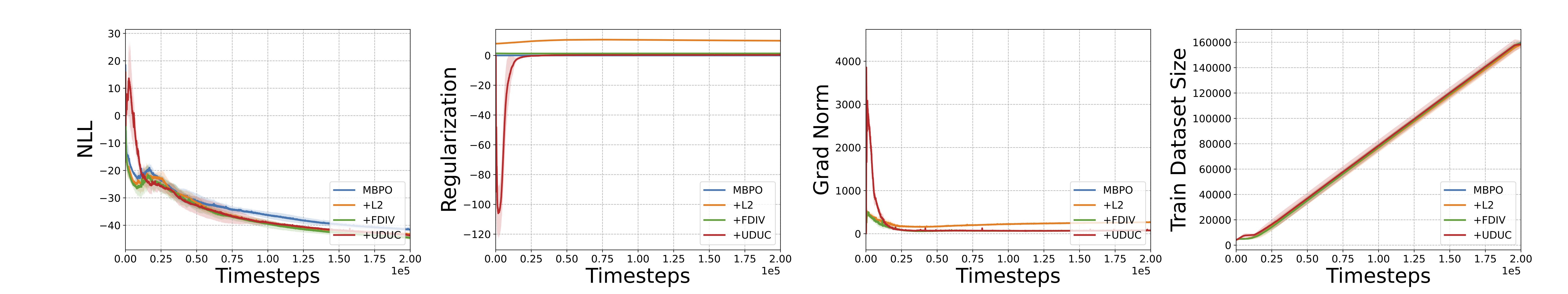}
    	    \caption{\texttt{walker\_stand}}
            \vspace{15pt}
    	\label{fig:walker_stand_train_additional}
        \end{subfigure}
        \begin{subfigure}[b]{1.0\linewidth}
            \centering        	    
            \includegraphics[width=\textwidth]{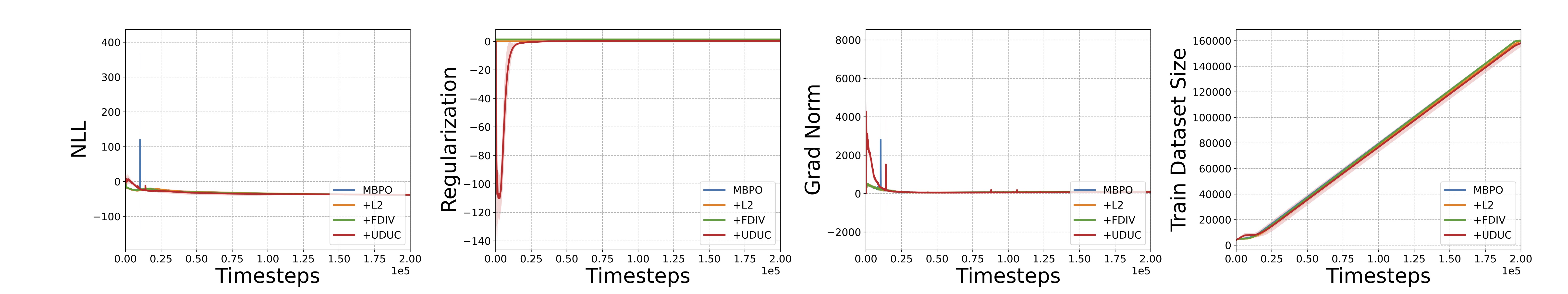}
            \caption{\texttt{walker\_walk}}
             \vspace{15pt}
    	\label{fig:walker_walk_train_additional}
        \end{subfigure}
        \begin{subfigure}[b]{1.0\linewidth}
    	\centering        	   
            \includegraphics[width=\textwidth]{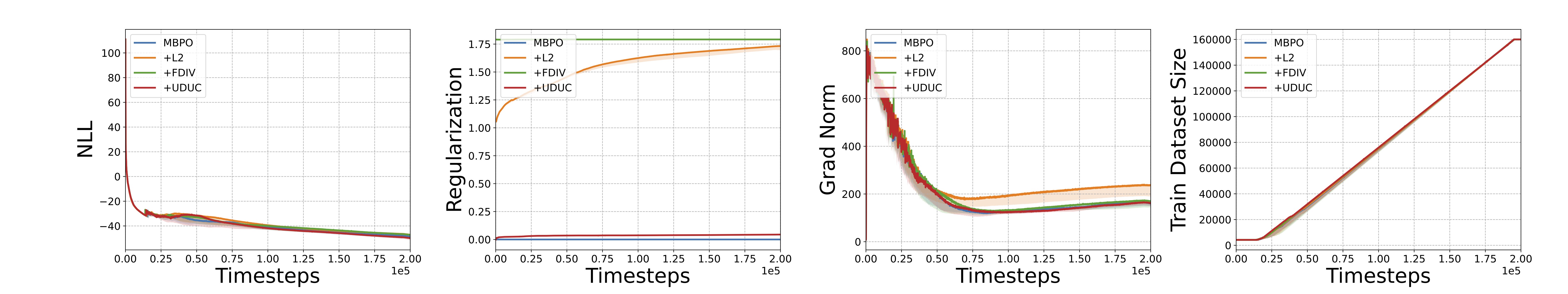}
    	    \caption{\texttt{humanoid\_walk}}
            \vspace{15pt}
    	\label{fig:humanoid_walk_train_additional}
        \end{subfigure}
    \caption{Additional training curves on robot control tasks. The x-axis is the environment time steps and the y-axis is neg log-likelihood, regularization, gradients norm and model training dataset. All graphs are plotted with median and 25\%-75\% percentile shading across 5 random seeds.}
    \vspace{15pt}
    \label{fig:rwrl_train_addtional}
    \end{figure*}


\end{document}